\documentclass[10pt,twocolumn,letterpaper]{article}

\usepackage[pagenumbers]{cvpr} 

\usepackage[dvipsnames]{xcolor}

\definecolor{cvprblue}{rgb}{0.21,0.49,0.74}
\usepackage[pagebackref,breaklinks,colorlinks,citecolor=cvprblue]{hyperref}
\usepackage{multirow}
\usepackage{xcolor}
\usepackage{colortbl}

\makeatletter
\newcommand{\crefordefault}[2]{%
  \@ifundefined{r@#1}{#2}{\cref{#1}}%
}
\makeatother
\newcommand{\blue}[1]{\textcolor{blue}{#1}}

\newif\ifshowcomments
\newcommand{\vicentedone}[1]{}
\newcommand{\alexdone}[1]{}

\showcommentstrue  
\ifshowcomments
     \newcommand{\vicente}[1]{{\color{magenta}[vicente: #1]}}
    \newcommand{\alex}[1]{{\color{red}[alex: #1]}}

\else
    \newcommand{\vicente}[1]{}
    \newcommand{\alex}[1]{}
\fi
\definecolor{darkgreen}{RGB}{0,112,0}

\title{Improved Visual Grounding through Self-Consistent Explanations}

\author{
Ruozhen He\textsuperscript{$1$}
\quad Paola Cascante-Bonilla\textsuperscript{$1$} 
\quad Ziyan Yang\textsuperscript{$1$} 
\quad Alexander C. Berg\textsuperscript{$2$} 
\quad Vicente Ordonez\textsuperscript{$1$}\\
\textsuperscript{$1$}Rice University 
\quad \textsuperscript{$2$}University of California, Irvine \\
{\tt\small catherine.he@rice.edu, pc51@rice.edu, zy47@rice.edu, bergac@uci.edu, vicenteor@rice.edu}
}

\begin{document}
\maketitle
\begin{abstract}
Vision-and-language models trained to match images with text can be combined with visual explanation methods to point to the locations of specific objects in an image. 
Our work shows that the localization --``grounding''-- abilities of these models can be further improved by finetuning for self-consistent visual explanations. 
We propose a strategy for augmenting existing text-image datasets with paraphrases using a large language model, and SelfEQ, a weakly-supervised strategy on visual explanation maps for paraphrases that encourages self-consistency. Specifically, for an input textual phrase, we attempt to generate a paraphrase and finetune the model so that the phrase and paraphrase map to the same region in the image.
We posit that this both expands the vocabulary that the model is able to handle, and improves the quality of the object locations highlighted by gradient-based visual explanation methods (e.g.~GradCAM). We demonstrate that SelfEQ improves performance on Flickr30k, ReferIt, and \mbox{RefCOCO+} over a strong baseline method and several prior works. Particularly, comparing to other methods that do not use any type of box annotations, we obtain 84.07\% on Flickr30k (an absolute improvement of 4.69\%), 67.40\% on ReferIt (an absolute improvement of 7.68\%), and 75.10\%, 55.49\% on RefCOCO+ test sets A and B respectively (an absolute improvement of 3.74\% on average).
\end{abstract}

\section{Introduction}
\label{sec:intro}

Vision-and-language models that are trained to associate images with text have shown to be effective for many tasks and benchmarks~\cite{clip,albef,jia2021scaling,li2022grounded}, including object detection~\cite{zareian2021open,gu2021open} and image segmentation~\cite{xu2022groupvit,luddecke2022image,ghiasi2022scaling}. Since these models are typically trained with in-the-wild data from the web, they can handle a wide range of vocabulary for objects as long as they are well represented in the training data. These models are often remarkably accurate~\cite{align,singh2022flava,blip,cyclip} even without tuning them to perform well in any particular downstream task~\cite{filip,declip,pyramidclip}. 
The ALBEF model~\cite{albef} was particularly capable of visual ``grounding'' -- or in other words -- the ability to localize objects in images by simply using it in conjunction with a visual explanation method such as GradCAM~\cite{selvaraju2017grad}. This capability is particularly remarkable given that this model was only supervised with images and text but no object location annotations of any type. 

\begin{figure}[t!]
\begin{center}
\vspace{0.05in}
\includegraphics[width=0.48\textwidth]{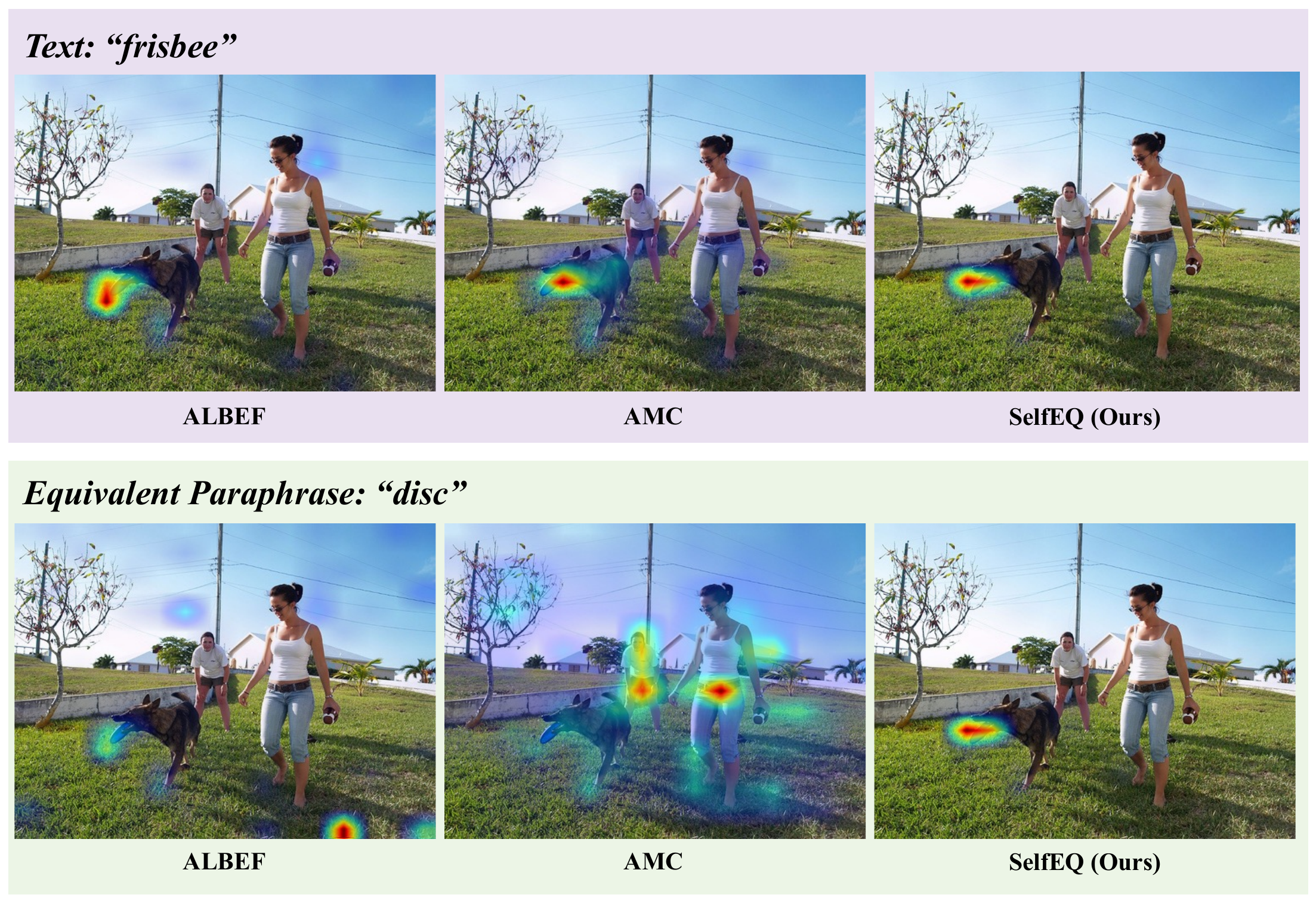}
\vspace{-0.2in}
\caption{
Previous models can localize the word \emph{frisbee}, but struggle with equivalent but more 
uncommon referents such as \emph{disc}. A model tuned with our proposed SelfEQ objective encourages consistent visual explanations for paraphrased prompts and performs well on both examples. SelfEQ not only enables a larger working vocabulary but also improves overall localization performance.
}
\label{fig:teaser}
\vspace{-0.3in}
\end{center}
\end{figure}

In order to improve the ability of vision-and-language models to perform localization, many methods have incorporated further finetuning with either box or segment annotations, or rely on pretrained object detectors or box proposal networks~\cite{chen2020uniter,dou2021improving,gupta2020contrastive,yang2023improving,glip,mdetr}. 
Our work instead aims to improve the localization capabilities of models trained only on image-text pairs through weak supervision. 
But, how can we improve the ability of a model to localize objects without access to object location annotations? 
Consider the example in Figure~\ref{fig:teaser} where a model is tasked with pointing to the location of the object \emph{frisbee} in this image. The baseline model succeeds at finding the object but is unsuccessful at locating the object when prompted with the equivalent but more generic name \emph{disc}. Regardless of the ability for the base model to find either of these, the visual explanations for these two prompts should be the same since the query refers to the very same object in both cases. Our work exploits this property by first generating paraphrases using a large language model and then proposing a weakly-supervised \textbf{Self}-consistency \textbf{EQ}uivalence Tuning (SelfEQ) objective that encourages consistent visual explanations between paraphrased input text pairs that refer to the same object or region in a given image.

Given a base pre-trained vision-and-language model purely trained on image-text pairs such as ALBEF~\cite{albef}, SelfEQ tunes the model so that for a given input image and text pair, the visual attention map extracted using GradCAM~\cite{selvaraju2017grad} produces a similar visual attention map when provided with the same image and a text paraphrase. Figure~\ref{fig:method} provides an overview of our method. Another contribution of our work consists in exploiting a large language model (LLM) to automatically generate paraphrases for existing datasets such as Visual Genome~\cite{krishna2017visual} that contains textual descriptions of individual objects and regions, or MS-COCO~\cite{lin2014microsoft} and CC3M~\cite{sharma2018conceptual} that contain global image descriptions. We find that SelfEQ not only expands the vocabulary of objects that the base model is able to localize but more importantly, improves the visual grounding capabilities of the model on standard benchmarks such as referring expression comprehension on the ReferIt benchmark~\cite{kazemzadeh2014referitgame} and region-phrase grounding in the Flickr30K Entities benchmark~\cite{plummer2015flickr30k}. 
In summary, our key contributions are as follows:

\begin{itemize}
\item We design a novel objective, SelfEQ, to encourage vision-and-language models to generate self-consistent visual explanations for equivalent text phrases, thereby improving grounding capabilities while expanding the working vocabulary of the model. 
\item We propose to prompt large language models for generating paraphrased image descriptions of individual objects or regions. Particularly, we adopt Vicuna-13B~\cite{vicuna2023} and design text prompts
to obtain high quality paraphrases.
\item We demonstrate the effectiveness of our method by outperforming previous methods, leading to $4.69\%$ improvement on Flickr30k, $7.68\%$ improvement on ReferIt, and $3.74\%$ improvement on RefCOCO+. 
\end{itemize}

Finally, we plan to release our code, generated paraphrases and model checkpoints upon publication.

\section{Related Work}
\label{sec:relatedWork}

Our work is related to previous methods on visual grounding, especially those that are trained under weak supervision understood as without the use of bounding box or segment annotations and relying only on image-text pairs. From a technical perspective our work is related to methods that optimize visual explanations to improve the underlying model. 

\vspace{0.05in}
\noindent
Visual grounding consists of localizing an input textual description in an image.  
Supervised methods are provided with text-image pairs and corresponding bounding boxes~\cite{deng2018visual, dou2021improving, deng2021transvg, yang2023improving,mdetr}. 
Other supervised methods leverage pretrained object detectors 
to obtain a region of interest and then identify the region that aligns most closely under their textual representations~\cite{datta2019align2ground,gupta2020contrastive,wang2019phrase,lu202012,gomel2023box,chen2020uniter}. In both cases, these methods use some form of box supervision during pre-training or at test time by relying on a pre-trained object detector. In contrast, our work focuses exclusively in the scenario where no bounding boxes or segment annotations are available at any stage.

\noindent
\textit{Weakly-Supervised Grounding.} Our setup is similar to that of Arbelle~et~al~\cite{arbelle2021detector} where no box annotations or object detectors are used for grounding. This work proposes Grounding by Separation (GbS) where a model is trained on randomly alpha-blended pairs of images and the goal is to separate them conditioned on text prompts. Our method instead relies on data augmentation on the text side and while our method shows favorable results, our contribution is orthogonal.
Shaharabany~et~al~\cite{shaharabany2022looking} builds a weakly-supervised phrase grounding model by creating a large amount of data by combining region boxes with a BLIP captioning model.
Later work by Shaharabany~and~Wolf~\cite{shaharabany2023similarity} 
employs layer-wise relevance propagation~\cite{montavon2017explaining} to integrate relevancy and gradient information with the scores computed from each attention head in the transformer layers~\cite{chefer2021transformer}, or residual connections~\cite{abnar2020quantifying}. Our work compares favorably or on par with these methods but since we rely on gradient-based explanations, our work does not require making any modifications to the base network.

\noindent
\textit{Visual Explanation Tuning.} Related to our method is the strategy used by \mbox{ALBEF}~\cite{albef} where the model is only supervised on image-text pairs and performs grounding through GradCAM~\cite{selvaraju2017grad}, a gradient-based visual explanation method that outputs a heatmap indicating spatial relevance. Earlier, Xiao~et~al~\cite{xiao2017weakly} used a similar strategy but further optimized gradient-based explanations using structural constraints derived from text.  Recently, AMC~\cite{yang2023improving} follows this strategy but further adds box supervision on the output maps using a margin-based loss. We adopt ALBEF~\cite{albef} as our base model and also adopt a gradient-based explanation strategy but unlike AMC~\cite{yang2023improving} we do not rely on box annotations for tuning this model and use our proposed SelfEQ objective instead. Javed~et~al~\cite{javed2018learning} proposed an objective function that encourages consistent representations in embedding space for the same input prompt on different images. In contrast SelfEQ encourages consistent visual explanations for different prompts on the same image by relying on automatically generated paraphrases. Akbari~et~al~\cite{akbari2019multi} also optimizes attention maps but in their formulation the model backbone is modified to explicitly incorporate attention instead of relying on gradient-based attention maps.

\section{Method}
\begin{figure}[t]
  \centering
   \includegraphics[width=\columnwidth]{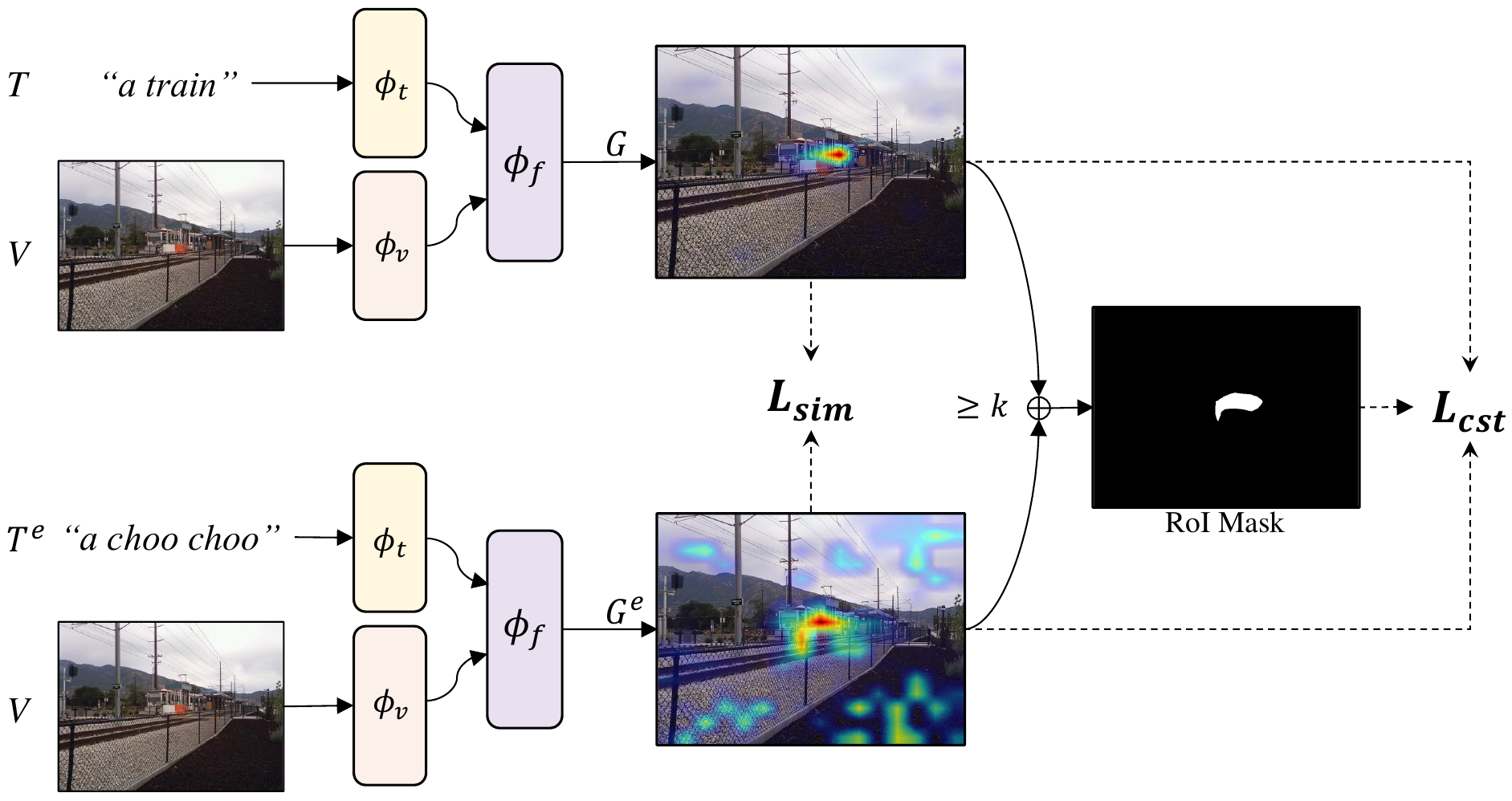}

   \caption{Overview of our proposed weakly-supervised \textbf{Self}-consistency \textbf{EQ}uivalence tuning objective. We input image-text and image-paraphrase pairs $\langle V, T\rangle$ and $\langle V, T^e\rangle$ to our base pre-trained vision-and-language model.
   We then obtain gradient-based visual explanations $\langle G,G^e \rangle$ and 
   compute a similarity loss 
   between them. We also define an overlapping region of interest mask and 
   encourage the model to predict consistently high saliency scores within this mask for each input pair.
   }
   \label{fig:method}
   \vspace{-0.1in}
\end{figure}

We start from a base vision-language model composed of a text decoder $\phi_t$
, an image encoder $\phi_v$ 
, and a multimodal fusion encoder $\phi_f$, along with a dataset $D$ to finetune this model consisting of image-text pairs $\langle T,V \rangle$.
Section~\ref{sec:loss}, introduces the training objectives for the base model which we also adopt for finetuning our baseline and are also used in conjunction with SelfEQ to finetune our final model.
Section~\ref{sec:selfconsistency} introduces in detail SelfEQ, our self-consistency equivalence tuning objective that assumes the existence of paraphrases $T^e$ for each input image-text pair $\langle T, V\rangle$, and Section~\ref{sec:data} describes our approach for automatically generating paraphrases $T^e$ using LLM-prompting. Figure~\ref{fig:method} presents an overview of our proposed method.

\subsection{Base Model: Preliminaries}
\label{sec:loss}

Our base vision-and-language model is ALBEF~\cite{albef} which relies on three widely used objectives for visual and textual representation learning: image-text matching, masked language modeling and a contrastive loss. We describe them here briefly as they are also re-used during fine-tuning.

\vspace{-0.2in}
\paragraph{Image-Text Matching Loss (ITM).} This loss is calculated using the output of the \texttt{[CLS]} token to predict if the input image and the input text are matching or not and is defined as follows: 
\begin{equation}
\small
\mathcal{L}_{\mathrm{itm}} = \;\mathbb{E}_{(V,T)\sim D}\;\mathcal{H}\left(\Vec{y}, \phi_f^{\mathrm{cls}}\left(\phi_v\left(V\right), \phi_t\left(T\right)\right)\right),
\end{equation}
where $\Vec{y}$ denotes a two-dimensional one-hot vector, indicating whether the sample $\langle V,T \rangle $ constitutes a match, $\phi_f^{\mathrm{cls}}$ represents a linear layer followed by a softmax function, and $\mathcal{H}$ is the cross entropy loss function.

\paragraph{Masking Language Modeling Loss (MLM).} 
This loss has been applied for various vision-language pretraining models~\cite{albef,chen2020uniter,li2019visualbert,lu2019vilbert}. It integrates the contextual text and the input image to infer masked words in the input text. After utilizing a linear layer and a softmax activation function $\phi_f^{\mathrm{m}}$ to individual output embeddings, the objective is expressed as:
\begin{equation}
\small
\mathcal{L}_{\mathrm{mlm}} = \;\mathbb{E}_{(V,T^\mathrm{-m})\sim D}\;\mathcal{H}\left(\Vec{t}^\mathrm{m}, \phi_f^{\mathrm{m}}\left(\phi_v\left(V\right), \phi_t\left(T^\mathrm{-m}\right)\right)\right),
\end{equation}
where the one-hot vector $\Vec{t}^\mathrm{m}$ denotes the masked token, and $T^\mathrm{-m}$ represents the input masked text.

\vspace{-0.1in}
\paragraph{Image-Text Contrastive Loss (ITC).} It improves the alignment between visual and textual representations by bringing closer the representations for corresponding text-image pairs relative to text-image pairs that do not correspond. 
This objective can be defined as follows:
\begin{equation}
\small
\begin{aligned}
\mathcal{L}_{\mathrm{itc}} = \mathbb{E}_{(V,T)\sim D} \frac{1}{2} \biggl[  \mathcal{H}\biggl(\Vec{y}, 
\frac{\mathrm{exp}\left(\phi_v(V) 
\cdot \phi_t(T)\right) / \tau}{\sum_{b=1}^{B} \mathrm{s}\left(V, T_b\right)}\biggr) \\
+ \mathcal{H}\biggl(\Vec{y}, 
\frac{\mathrm{exp}\left(\phi_t(T) 
\cdot \phi_v(V)\right) / \tau}{\sum_{b=1}^{B} \mathrm{s}\left(T, V_b\right)} \biggr) \biggr],
\end{aligned}
\end{equation}
where $B$ is the number of negative sample pairs, and $\tau$ is a temperature parameter for the softmax function.

The training objective for the base model is a combination of the previous three loss functions:
\begin{equation}
\small
    \mathcal{L}_{\mathrm{vl}}=\mathcal{L}_{\mathrm{itm}}+\mathcal{L}_{\mathrm{mlm}}+\mathcal{L}_{\mathrm{itc}}.
\end{equation}
This loss $\mathcal{L}_{\mathrm{vl}}$ will also be used to tune our baseline model.

\subsection{Self-Consistency Equivalence Tuning}
\label{sec:selfconsistency}
SelfEQ assumes that the model has access to paraphrases $T^e$ for each input image-text pair $\langle V,T\rangle$ or, in practice, for a subset of those samples. Therefore we assume a finetuning dataset $D'$ with triplets $\langle V, T, T^e \rangle$ such that $T_e$ exists for a corresponding input text $T$.
The first step to define our SelfEQ objective is to generate the explanation heatmaps (\textit{i.e.,}~attention maps) through GradCAM~\cite{selvaraju2017grad} conditioned on the input text. We extract intermediate feature maps from the multimodal interactive encoder $\phi_f$ for input pairs $\langle V,T\rangle$ and $\langle V, T^e \rangle$ as follows: 
\begin{equation}
\small
    F = \phi\left(\phi_v(V), \phi_t(T)\right),
    F^e = \phi\left(\phi_v(V), \phi_t(T^e)\right),
\end{equation}
where $\phi$ denotes the feature map extraction operation. We then proceed to calculate the gradient of $F$ and $F^e$ related to the image-text matching score $\mathcal{L}_{\mathrm{itm}}$. This computation yields the attention maps for the original text and paraphrased text, referred to as $G$ and $G^e$, respectively:
\begin{equation}
\small
\begin{aligned}
    G &= \mathrm{ReLU}\left(F \odot \triangledown \mathcal{H} \left(\Vec{y}, \phi_f^{cls} \left(\phi_v(V), \phi_t(T)\right)\right)\right), \\
    G^e &= \mathrm{ReLU}\left(F^e \odot \triangledown \mathcal{H} \left(\Vec{y}, \phi_f^{cls} \left(\phi_v(V), \phi_t(T^e)\right)\right)\right).
\end{aligned}
\end{equation}

Our SelfEQ tuning is based on the premise that if a vision-language model is identified as self-consistent, the attention maps produced for both the text and its equivalent paraphrase should yield nearly identical results. To achieve this, 
we first apply a simple mean squared error loss over the produced heatmaps so that their $\ell_2$ distance is minimized and thus become more similar. 
\begin{equation}
\small
    \mathcal{L}_{\mathrm{sim}} = \\
    \mathbb{E}_{(V,T,T^e)\sim D'}\biggl[\frac{1}{N} \sum_{i,j}(G_{i,j} - G^e_{i,j})^2\biggr].
\end{equation}

Nevertheless, while minimizing a sum of pixel-wise distances contributes to self-consistency, without a regularization term this loss can easily fall into a trivial solution. For instance, it could lead to attention maps with uniformly negative or positive predictions, or just really small values. To address this limitation, we propose to integrate these heatmaps by defining a Region of Interest (RoI) mask. This mask is designed to preserve regions within the attention maps that possibly contain correct predictions. Our approach hinges on the observation that, despite the predictions of equivalent textual inputs being inconsistent, sometimes regions with large values or regions that overlap between the two heatmaps tend to be correct. As such, we assume that if the sum of attention scores at a given position $(i,j)$ exceeds a certain threshold $k$, it is likely indicative of a correct prediction. We formalize the condition as follows:
\begin{equation}
\small
    M_{i,j} =
    \begin{cases}
      1, (G_{i,j} + G^e_{i,j}) \geq k \\
      0, (G_{i,j} + G^e_{i,j}) < k
    \end{cases}
    .
\end{equation}
The attention maps within RoI masks are obtained by element-wise multiplication as follows:
\begin{equation}
\small
    R = G \odot M, \quad R^e = G^e \odot M.
\end{equation}

The integration of RoI masks allows us to use equivalent texts for mutual supervision, refining and improving the accuracy and providing regularization for the previously defined distance-based loss. Moreover, it could potentially address errors owing to unknown or less common words by vocabulary expansion. Presuming one of the textual expressions is known and correctly understood, the model could extrapolate the meaning of the other equivalent expression via weak supervision. To implement it, we first compute  the mean $\mu_{RoI}, \mu_{RoI}^e$ and the standard deviation $\sigma_{RoI}, \sigma_{RoI}^e$ within the RoI as follows:

\begin{equation}
\small
    \mu_{RoI}= \frac{\sum_{i,j} R_{i,j}}{\sum_{i,j} M_{i,j}},
    \mu_{RoI}^e= \frac{\sum_{i,j} R^{e}_{i,j}}{\sum_{i,j} M_{i,j}},
\end{equation}

\begin{equation}
\small
\begin{aligned}
\sigma_{RoI} &= \sqrt{\frac{\sum_{i,j} M_{i,j} \cdot (R_{i,j} - \mu_{RoI})^2 }{\sum_{i,j} M_{i,j}}}, \\
\sigma_{RoI}^e &= \sqrt{\frac{\sum_{i,j} M_{i,j} \cdot (R^e_{i,j} - \mu_{RoI}^e)^2 }{\sum_{i,j} M_{i,j}}}.
\end{aligned}
\end{equation}

We propose a consistency loss ($\mathcal{L}_{\mathrm{cst}}$), expecting the RoI regions of attention maps to achieve consistently high scores, further reinforcing self-consistency, accuracy, and potential vocabulary expansion. This objective is formulated as follows:
\begin{equation}
\small
\begin{aligned}
    \mathcal{L}_{\mathrm{cst}} = & \mathbb{E}_{(V,T,T^e)\sim D'}  \biggl[\sigma_{RoI} + \sigma_{RoI}^e + \\
    &\mathrm{max}(0, {k}/{2}-\mu_{RoI}) + \mathrm{max}(0, {k}/{2}-\mu_{RoI}^{e})  \biggr].
\end{aligned}
\end{equation}

Finally, the objective of our self-consistency equivalence tuning is expressed as:

\begin{equation}
\small
    \mathcal{L}_{\mathrm{SelfEQ}} =  \mathcal{L}_{\mathrm{sim}} + \lambda \cdot \mathcal{L}_{\mathrm{cst}},
\end{equation}
where $\lambda$ is a hyperparameter to control the relative influence of each loss. 

 \subsection{Self-Consistency Data Augmentation}
\label{sec:data}

In this section we define a function $\mathcal{F}$ that can automatically map input textual phrases $T$ as paraphrases $T^e$ without the need to rely on human annotations such that $T^e \sim \mathcal{F}(T)$. We achieve this goal through a two-level prompting approach using a large language model which we describe in detail below.

\vspace{0.05in}
\noindent
\textbf{Phrase Chunking: } Using our first-level prompts, we aim to augment the original text using phrase chunking to encourage global captions to concentrate on more specific regions. 
Visual grounding seeks to localize objects in images based on textual inputs. In contrast, global captions usually describe the entire image, typically describing several objects. While training on global captions could be beneficial for learning cross-modal information, it may lead the model to predict a broader region (\textit{i.e.}, global context) rather than a specific region. 
Phrase chunking (\textit{i.e.}, shallow parsing~\cite{zhai2017neural}) aims to identify continuous sequences of tokens representing syntactic units, enabling the extraction of phrases from unstructured text. 
We leverage an LLM to segment global captions into object-centric short phrases. 
During training, we use these image-chunk pairs instead of global captions, effectively guiding the model attention toward localized regions of interest.
We \emph{refer the reader to the supplementary material for prompting details and generated examples.}

\vspace{0.05in}
\noindent
\textbf{Paraphrase Generation: } 
Our SelfEQ approach involves feeding the model with pairs of textual descriptions that refer to the same underlying concept, with the expectation that the model can make similar predictions for these equivalent description pairs $\langle T, T^e \rangle$. We augment our dataset by transforming the region-based captions (\textit{i.e.}, text that only refers to a region in the image) and the object-centric short phrases we obtained from phrase chunking into equivalent paraphrases referring to the same concept through our second-level LLM-prompts. 

There are many ways to paraphrase, including substituting words, altering sentence structures, and rewriting sentences based on semantics. However, considering that self-consistency in vision and language is relatively under-explored, we adopt a straightforward strategy: Replacing the primary object in the sentence while retaining all other attributes. This strategy yields several benefits. First, it provides a consistent context, which serves as a reference for the model to identify equivalent descriptions. This enables the equivalent relationships of paraphrases to be learned intuitively. Second, it simplifies prompt designing and post-processing by detecting the primary object and generating its synonym. 

\begin{figure}[t]
  \centering
   \includegraphics[width=\columnwidth]{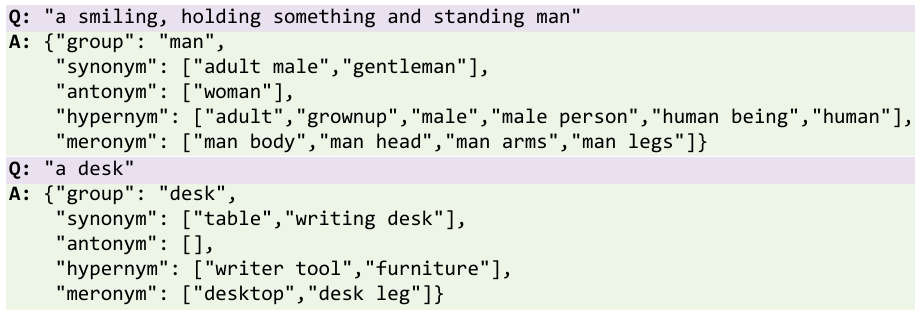}
   \vspace{-0.2in}
   \caption{
   Two samples from our LLM-prompt for paraphrase generation. The first set showcases an example of a region-based caption, and the second set shows a non-sentence phrase. \textbf{Q} is the query text and \textbf{A} is the expected answer.
   }
   \label{fig:prompt_sample}
   \vspace{-0.2in}
\end{figure}

To generate paraphrases for the dataset consisting of region-based captions, we select four textual descriptions 
in which the primary noun plays different syntactic roles. 
We further select two non-sentence phrases 
as examples of query texts in our prompts.
We show an example of a region-based caption and a non-sentence phrase in Figure~\ref{fig:prompt_sample}.
To design our prompt, we identify the primary object in the query text \textbf{Q}. Then we use WordNet~\cite{miller1995wordnet} to obtain synonyms automatically and further remove inaccurate or invalid words. 
We add \textbf{A} to indicate the expected answer and include other relationships such as \textit{antonym, hypernym} and \textit{meronym} to provide richer contexts for LLM in-context learning.
\emph{Additional prompting details and paraphrase samples are provided in the supplementary material.}

This two-level prompt-based LLM augmentation approach ensures that our model is exposed to textual inputs that share the same concept while varying in linguistic representation, thereby promoting self-consistency and working vocabulary expansion.

\newcommand{\grayfont}{\color{black!55}}
\newcommand{\violetcell}[1]{\cellcolor{violet!8}#1}

\begin{table}[t!]
\centering
\small
\renewcommand{\arraystretch}{1.2}
\begin{tabular}{llccccc}
\toprule
&\textbf{Method} & 
\textbf{Training} & \textbf{Flickr30k} & \textbf{ReferIt} \\
\midrule

\multirow{5}{*}{\rotatebox[origin=c]{90}{{\grayfont\footnotesize Box Supervision}}}

&\grayfont Align2Ground~\cite{datta2019align2ground} 
& 
\grayfont VG-boxes & \grayfont 71.00 &\grayfont - \\

&\grayfont 12-in-1 \cite{lu202012} 
& \grayfont VG-boxes &\grayfont 76.40 &\grayfont - \\

&\grayfont InfoGround \cite{gupta2020contrastive} 
& \grayfont VG-boxes & \grayfont76.74  &\grayfont - \\

&\grayfont VMRM \cite{dou2021improving} 
& \grayfont VG-boxes & \grayfont 81.11  & - \\

&\grayfont AMC \cite{yang2023improving} 
& \grayfont VG-boxes & \grayfont \textbf{86.59} & \grayfont\textbf{73.17} \\

\midrule

\multirow{15}{*}{\rotatebox[origin=c]{90}{{\footnotesize Without Box Supervision}}}
&TD \cite{zhang2018top} 
& VG & 42.40 & 31.97 \\

&SSS \cite{javed2018learning} 
& VG & 49.10 & 39.98 \\

&MG-BiLSTM \cite{akbari2019multi} 
& VG & 57.91 & 62.76 \\

&MG-ELMo \cite{akbari2019multi} 
& VG & 60.08 & 60.01 \\

&GbS \cite{arbelle2021detector} 
& VG & 73.39 & 62.24 \\

&g \cite{shaharabany2022looking} 
& VG & 75.63 & 65.95 \\
&g++ \cite{shaharabany2023similarity} 
& VG & 79.95 & \textbf{70.25} \\

&\violetcell{$\mathrm{SelfEQ}$ (ours)} 
& \violetcell{VG} & \violetcell{\textbf{81.90}} & \violetcell{67.40} \\

\cmidrule{2-5}

&FCVC \cite{fang2015captions} 
& MS-COCO & 29.03 & 33.52 \\

&MG-BiLSTM \cite{akbari2019multi} 
& MS-COCO & 53.29 & 47.89 \\

&MG-ELMo \cite{akbari2019multi} 
& MS-COCO & 61.66 & 47.52 \\

&GbS \cite{arbelle2021detector} 
& MS-COCO & 74.50 & 49.26 \\

&g \cite{shaharabany2022looking} 
& MS-COCO & 75.43 & 61.03 \\

&g++ \cite{shaharabany2023similarity} 
& MS-COCO & 78.10 & 61.53 \\
&\violetcell{$\mathrm{SelfEQ}$ (ours)} 
& \violetcell{MS-COCO} & \violetcell{\textbf{84.07}} & \violetcell{\textbf{62.75}} \\
\bottomrule
\end{tabular}
\caption{Visual Grounding results on two benchmarks using pointing game accuracy with two training datasets. 
SelfEQ yields generally the best overall performance among weakly-supervised methods, and comes second to g++ on the ReferIt benchmark when trained using VG.
We also show at the top the results of methods using additional box supervision from Visual Genome (VG) either directly or through an object detector.
\vspace{-0.15in}
}
\label{tab:result}
\end{table}

\section{Experimental Settings}

\paragraph{Training.}
We use ALBEF~\cite{albef} as our base model in all our experiments, given its reported off-the-shelf visual grounding performance
via GradCAM~\cite{selvaraju2017grad}. ALBEF combines a ViT-B~\cite{dosovitskiy2020image} model for encoding images and a BERTbase~\cite{devlin2018bert} model for encoding text. It is pre-trained on a range of datasets, including ImageNet-1K~\cite{russakovsky2015imagenet}, Conceptual Captions~\cite{sharma2018conceptual}, SBU Captions~\cite{ordonez2011im2text}, MS-COCO~\cite{lin2014microsoft}, and Visual Genome (VG)~\cite{krishna2017visual} excluding box annotations. We finetune ALBEF with image-text pairs from 
VG and MS-COCO without any type of box supervision (i.e., no bounding boxes or object detectors), following prior work~\cite{akbari2019multi}. 
We further leverage Vicuna-13B~\cite{vicuna2023} as our LLM-prompting model to generate the object-centric short phrases (via shallow parsing or chunking) and the equivalent paraphrases for our self-consistency data augmentation.
Additionally, we validate the effectiveness of our SelfEQ tuning and self-consistency data augmentation method by training on a preprocessed subset of the Conceptual Captions 3M (CC3M) dataset~\cite{sharma2018conceptual}, which contains 
many noisy or unaligned web-crawled AltText-image pairs. 
With this subset, we achieve an absolute improvement of 2.15\% on Flickr30k, 3.32\% on ReferIt, and 1.33\% on RefCOCO+;
\emph{refer to the supplementary material for detailed CC3M experiments.}

\vspace{-0.2in}
\paragraph{Evaluation.}
We conduct evaluations using Flickr30k \cite{plummer2015flickr30k} and ReferIt \cite{kazemzadeh2014referitgame} under pointing game accuracy following previous weakly-supervised visual grounding works \cite{akbari2019multi, arbelle2021detector}. 
To underscore the competitive edge of our method, we also present its performance on RefCOCO+ \cite{yu2016modeling}, a challenging benchmark more commonly used for testing box-supervised methods~\cite{datta2019align2ground, lu202012, dou2021improving, gupta2020contrastive, yang2023improving}.

\subsection{Implementation Details}
Our experiments are conducted on a single computing node with 8 NVIDIA A40 GPUs. During the training phase, input images are resized to $256 \times 256$ and augmented with horizontal flipping, color jittering, and random grayscale conversion. We set up an Adam optimizer \cite{kingma2014adam} with a learning rate of 1e-5 and a batch size of 448 across all experiments. 
We empirically set the RoI threshold \(k\) to 0.8 and the hyperparameter $\lambda$ to 1.0. For training with raw image-text pairs from the datasets, we employ the vision-language objective \(\mathcal{L}_{\mathrm{vl}}\) (Sec. \ref{sec:loss}), while for the subset with equivalent paraphrases, we use our self-consistency equivalence tuning objective \(\mathcal{L}_{\mathrm{SelfEQ}}\) (Sec. \ref{sec:selfconsistency}) and corresponding vision-language objective \(\mathcal{L}_{\mathrm{vl}}^{e}\). The composite objective function is given by \(\mathcal{L} = \alpha \cdot \mathcal{L}_{\mathrm{vl}} + (1-\alpha) \cdot (\mathcal{L}_{\mathrm{SelfEQ}} + \mathcal{L}_{\mathrm{vl}}^{e})\), where \(\alpha\) is initially set to 0 and increments to 1, remaining constant after the second epoch. Our hyperparameter values and schedules were determined empirically on a small validation subset.

\begin{table}[t!]
\centering
\renewcommand{\arraystretch}{1.2}
\small
\begin{tabular}{lccc}
\toprule
\multirow{2}{*}{\textbf{Method}} & \multirow{2}{*}{\textbf{Box Supervision}} & \multicolumn{2}{c}{\textbf{RefCOCO+}} \\
\cmidrule(lr){3-4}
 & & \textbf{Test A} & \textbf{Test B} \\
\midrule

\grayfont InfoGround \cite{gupta2020contrastive} &\grayfont Yes &\grayfont 39.80 &\grayfont 41.11 \\

\grayfont VMRM \cite{dou2021improving} & \grayfont Yes &\grayfont 58.87 &\grayfont 50.32 \\
\grayfont AMC \cite{yang2023improving} & \grayfont Yes &\grayfont  \textbf{80.34} &\grayfont \textbf{64.55} \\
\midrule
ALBEF \cite{albef} & No & 69.35 & 53.77 \\
\violetcell $\mathrm{SelfEQ}$ (ours) &\violetcell No & \violetcell \textbf{75.10} & \violetcell \textbf{55.49} \\
\bottomrule
\end{tabular}
\caption{Results on RefCOCO+ pointing game
accuracy. SelfEQ shows significant improvements over off-the-shelf ALBEF and competitive results compared to box-supervised methods.
\vspace{-0.15in}
}
\label{tab:refcoco}
\end{table}

\section{Experimental Results}

Our resulting model obtains the best performance on the task of weakly-supervised visual grounding compared to most methods under this setting and is comparable to several prior works that rely on some of box supervision. Moreover, our qualitative results show that our method can handle paraphrases and a larger vocabulary without the needed to increment the training dataset significantly.

\vspace{0.05in}
\noindent
\textbf{Flickr30k and ReferIt.}
We evaluate the effectiveness of our proposed SelfEQ method in Table \ref{tab:result}, demonstrating its substantial lead over GradCAM-based weakly-supervised approaches. Our self-consistency equivalent tuning adapts well for both region-based (\textit{i.e.,}, VG)  and global-based (\textit{i.e.,}, COCO) image-text pairs, yielding a performance gain of 4.69\% on Flickr30k and 7.68\% on ReferIt, compared to our base model ALBEF (see first row in Table \ref{tab:extra_data}). Notably, our method outperforms almost all box-supervised methods on Flickr30k~\cite{datta2019align2ground, lu202012, gupta2020contrastive, dou2021improving}. In the weakly-supervised setting, our method only comes second on ReferIt compared to g++\cite{shaharabany2023similarity} when trained on Visual Genome image-text pairs. This method leverages a custom architecture to produce a mask and uses heatmap supervision from a CLIP~\cite{clip} model as pseudo-labels during training. We posit that our contribution is orthogonal and our approach would likely also benefit from similar supervision, since CLIP is trained in a much larger image-text pair dataset.  Despite differences, our method still obtains higher performance when trained on MS-COCO and the best performance compared to all weakly-supervised methods on Flickr30K region-phrase grounding.

\begin{figure}[t!] 
    \centering
    \includegraphics[width=\linewidth]{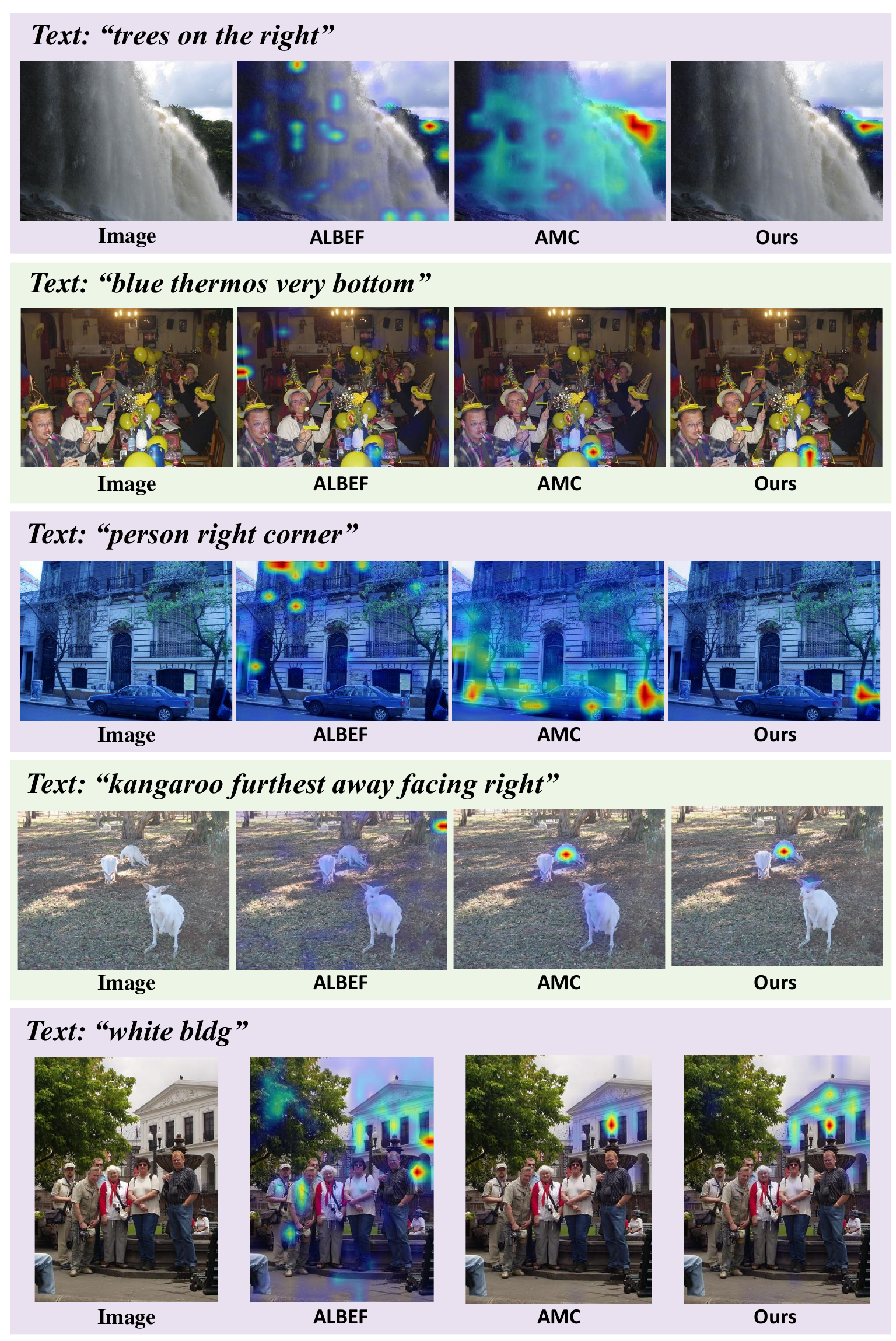}
    \vspace{-0.25in}
    \caption{Qualitative results of our method in challenging visual grounding scenarios compared to prior works. On top of each row we show the reference text, the first column shows the image, then we show our base model ALBEF, the SotA box-supervised method AMC, and finally we show results with our method SelfEQ. }
    \label{fig:comparison_grounding}
    \vspace{-0.15in}
\end{figure}

\vspace{0.05in}
\noindent
\textbf{RefCOCO+.} 
RefCOCO+~\cite{yu2016modeling} serves as a rigorous benchmark for visual grounding, typically used to evaluate box-supervised techniques. In Table \ref{tab:refcoco}, we present the performance of our weakly-supervised method (VG trained) against box-supervised methods. Our results indicate that our approach is competitive without reliance on any form of box annotations and significantly improves over the base ALBEF model.

\vspace{0.05in}
\noindent
\textbf{Visual Grounding Analysis.} 
Figure \ref{fig:comparison_grounding} provides qualitative results of our method in challenging scenarios, including occluded objects (row 1), small objects within complex scenes (row 2), objects partially shown in the corner of the image (row 3), multiple similar objects (row 4), and abbreviated text inputs (row 5). Our self-consistency equivalency tuning approach exhibits substantial improvements in the grounding capability of the base ALBEF~\cite{albef} model. Remarkably, our approach even outperforms the state-of-the-art box-supervised method AMC~\cite{yang2023improving} in multiple scenarios.

\begin{figure}[t!] 
    \centering
    \includegraphics[width=\linewidth]{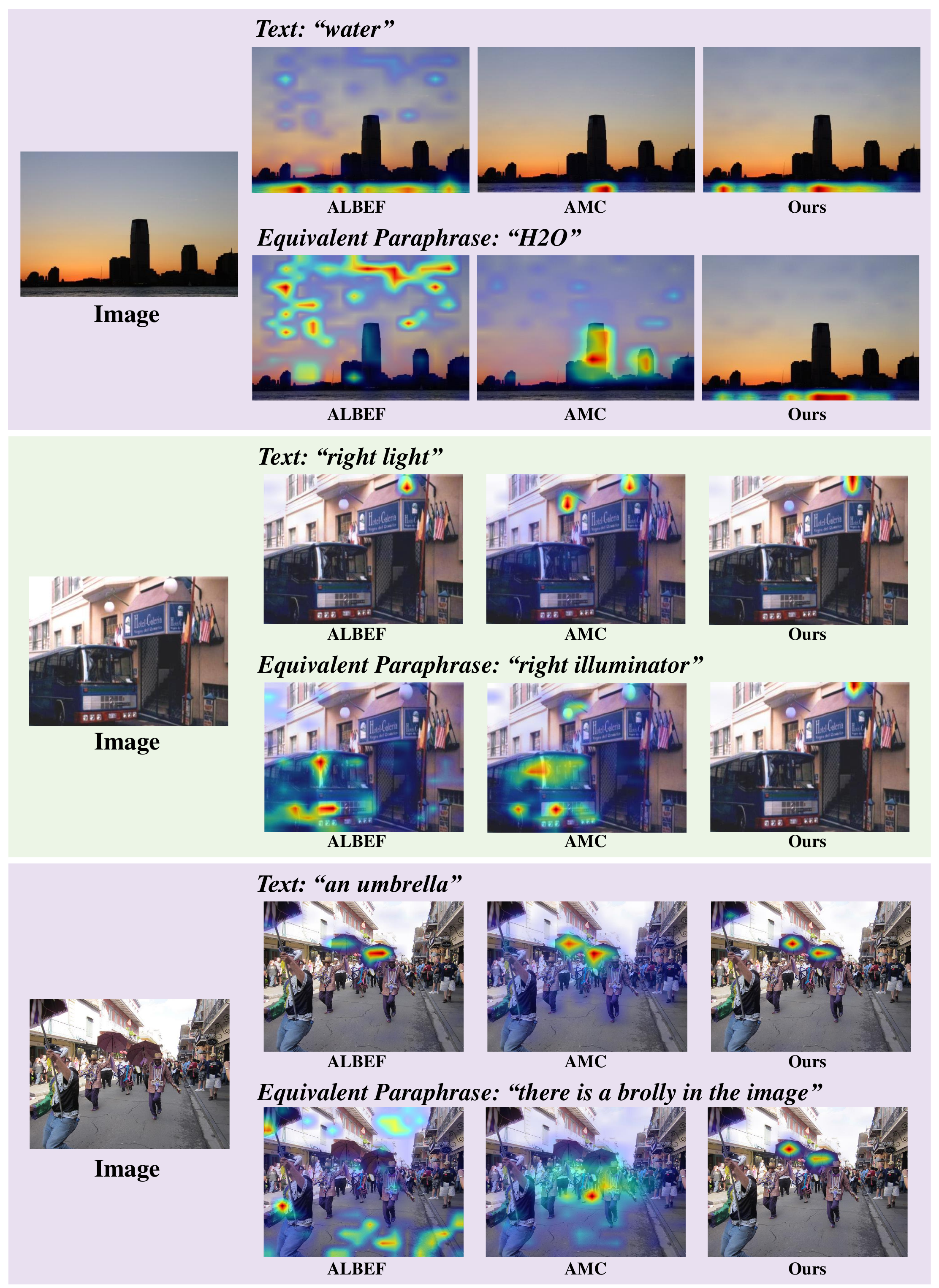}
    \vspace{-0.2in}
    \caption{Qualitative results of self-consistency across different equivalent paraphrases among different methods. For each image, we show a text caption referring to an object in the first row and an equivalent paraphrase in the second row. Each column shows the results of our base model ALBEF, the SotA box-supervised method AMC, and our SelfEQ method.}
    \label{fig:comparison_consistency}
    \vspace{-0.2in}
\end{figure}

\vspace{0.05in}
\noindent
\textbf{Self-Consistency Analysis.}
Figure~\ref{fig:comparison_consistency} demonstrates qualitative results for the self-consistency capability across different equivalent paraphrases, encompassing terminology (row 1), synonym substitutions (row 2), and regional slang combining with different sentence structures (row 3). Although other methods succeed in localizing certain phrases, they demonstrate inconsistencies for the equivalent paraphrases.
In contrast, our model finetuned with SelfEQ effectively establishes connections between semantically equivalent paraphrases, thereby enhancing the model self-consistency ability and potentially expanding its vocabulary.

\begin{table}[t!]
\centering
\small
\setlength{\tabcolsep}{4pt}
\begin{tabular}{lccccc}
\toprule
\multirow{2}{*}{\textbf{Data}} & \multirow{2}{*}{\textbf{Objective}} &  \multicolumn{2}{c}{\textbf{RefCOCO+}} & \multirow{2}{*}{\textbf{Flickr30k}} & \multirow{2}{*}{\textbf{ReferIt}} \\
\cmidrule(lr){3-4}
 & & \textbf{Test A} & \textbf{Test B} \\
\midrule
- & $\mathcal{L}_{\mathrm{vl}}$ & 69.35 & 53.77 & 79.38 & 59.72 \\
\midrule
$T$ & $\mathcal{L}_{\mathrm{vl}}$ & 72.30 & 54.22 & 78.75 & 65.86 \\
$T$ + $T^e$ & $\mathcal{L}_{\mathrm{vl}}$ & 71.55 & 53.51 & 78.05 & 64.57 \\
$T$ + $T^e$ & $\mathcal{L}_{\mathrm{SelfEQ}}$ & \textbf{75.10} & \textbf{55.49} & \textbf{81.90} & \textbf{67.40} \\
\bottomrule
\end{tabular}
\caption{Ablation studies on different ways to utilize extra equivalent paraphrased data. The first row is off-the-shelf ALBEF performance before tuning. $T$ denotes the textual captions from the dataset, and $T^e$ corresponds to the associated equivalent paraphrases. $\mathcal{L}_{\mathrm{vl}}$ is the vision-language objective, and $\mathcal{L}_{\mathrm{SelfEQ}}$ is our self-consistency equivalence tuning objective.
}
\label{tab:extra_data}
\vspace{-0.15in}
\end{table}

\subsection{Ablation Studies}

\begin{figure*}[th!] 
    \centering
    \includegraphics[width=0.24\textwidth]{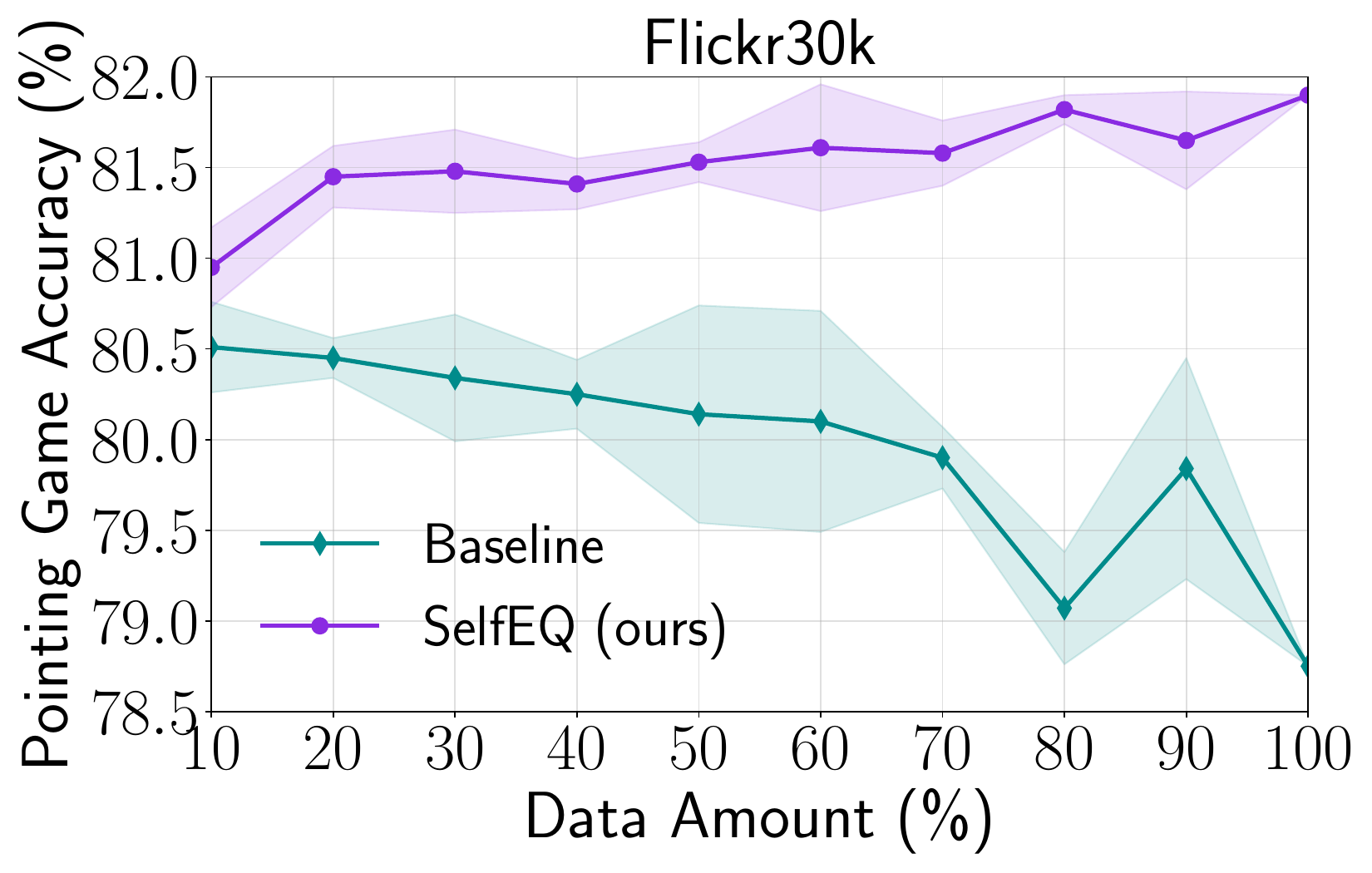}
    \includegraphics[width=0.24\textwidth]{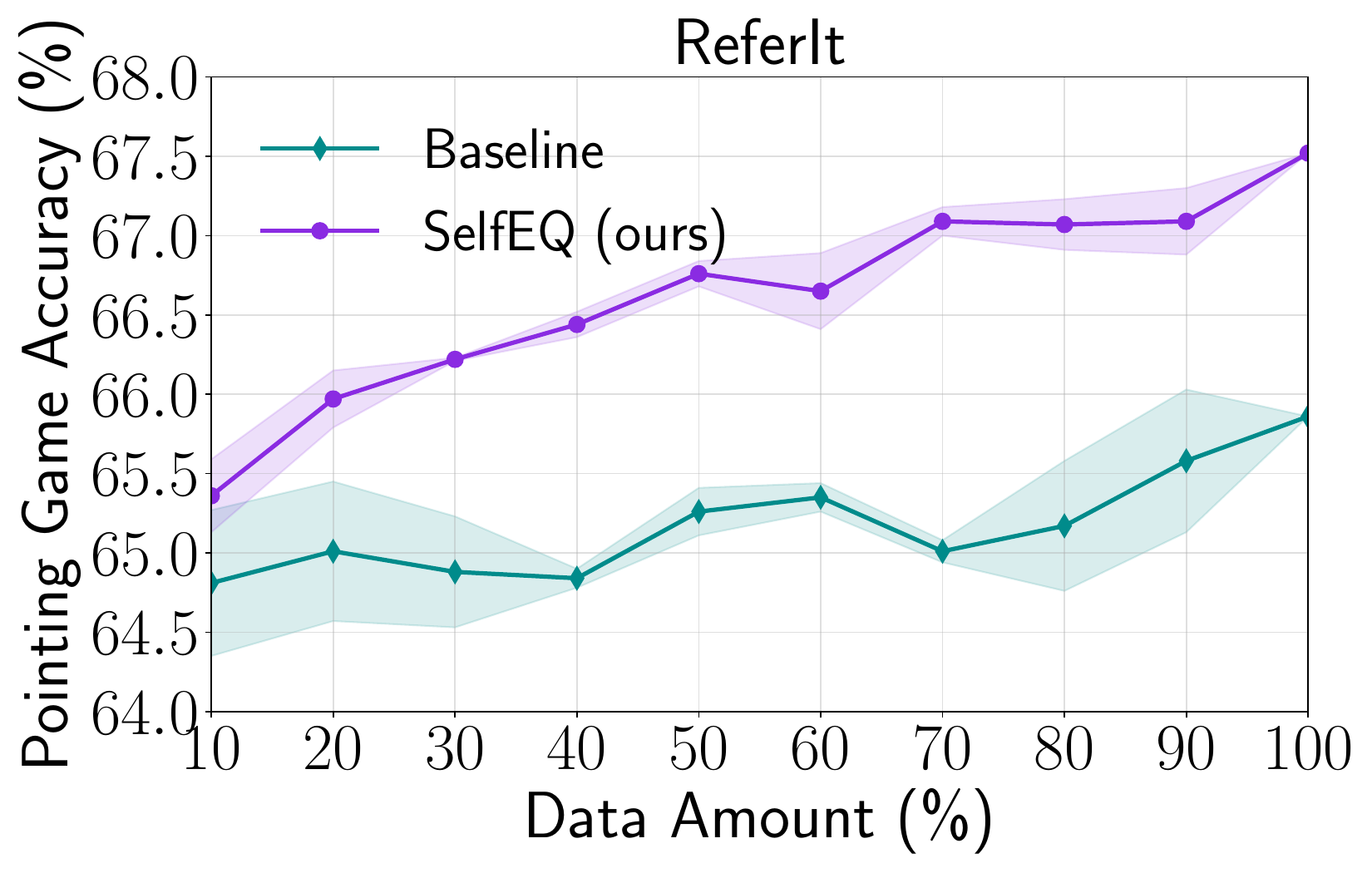}
    \includegraphics[width=0.24\textwidth]{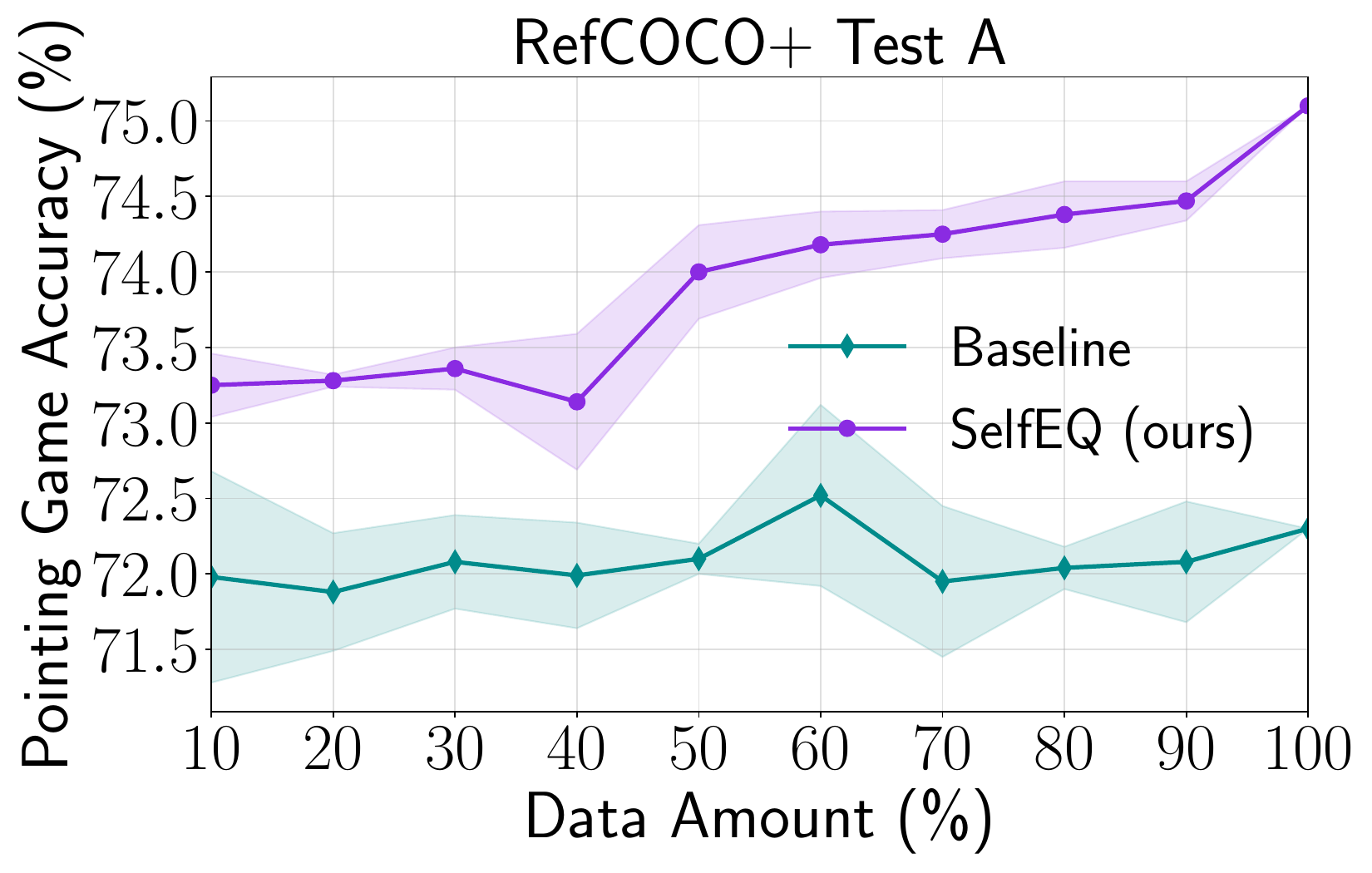}
    \includegraphics[width=0.24\textwidth]{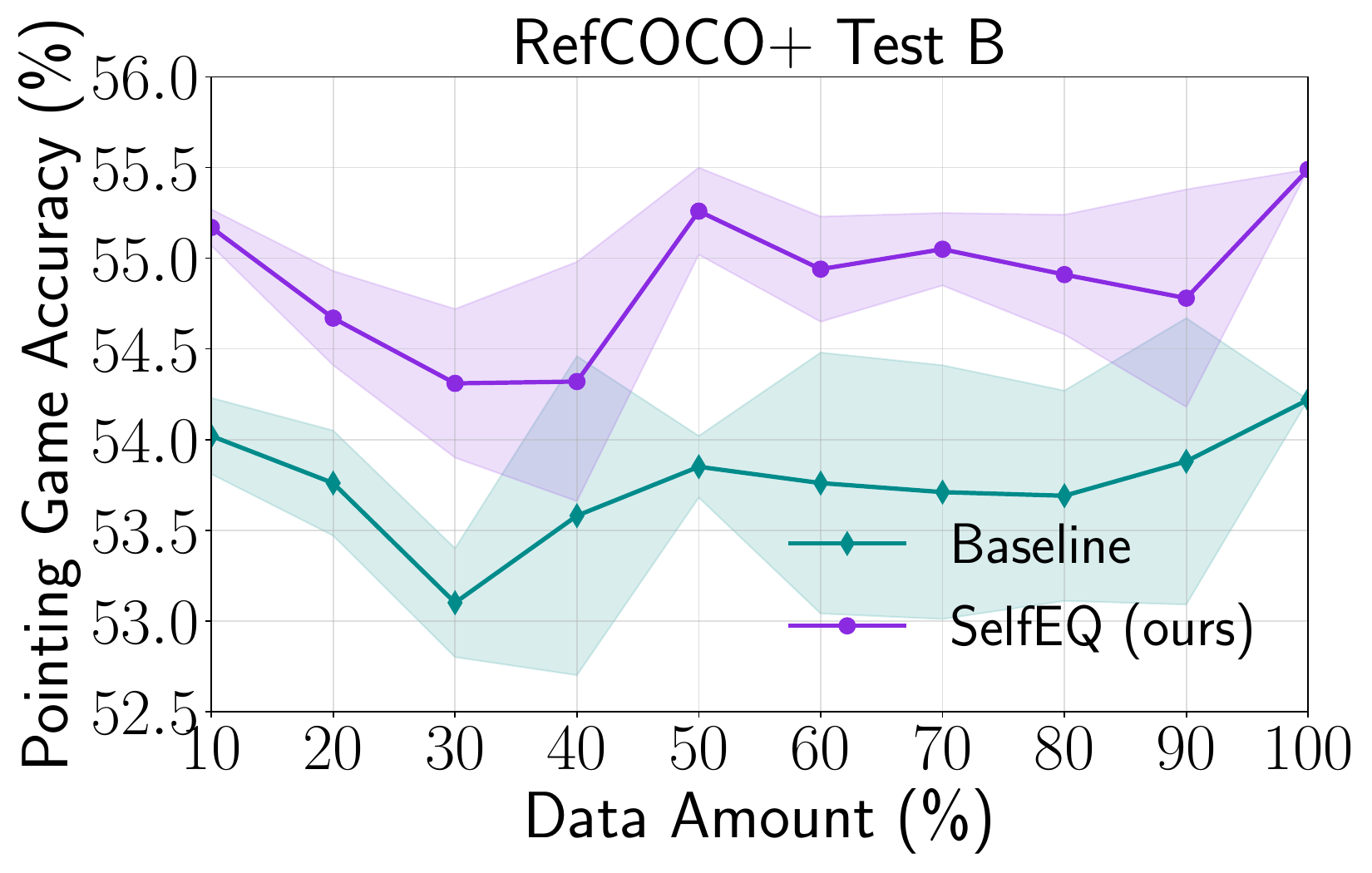}
    \vspace{-0.1in}
    \caption{Tuning performance with different data quantities on Flickr30k, ReferIt, RefCOCO+ Test A and Test B. 
    The purple and cyan lines represent SelfEQ (ours) and ALBEF baseline losses (vision-language objective), respectively.
    We show the impact of progressively augmenting captions via LLM-prompting for generating equivalent paraphrases tuned with our SelfEQ objective.
    Best viewed in color.
    } 
    \vspace{-0.1in}
    \label{fig:vg_amount}
\end{figure*}

\vspace{0.05in}
\noindent
\textbf{Data Quantity.} 
We conduct experiments to investigate the impact of the data quantity of our generated equivalent paraphrases and compare our self-consistency equivalence tuning strategy against vision-language objectives.
We randomly sample portions of data from VG by 10\% associated with our augmented equivalent paraphrases three times. 
To compare, we use the vision-language objective with VG text-image pairs as baselines.
In Figure~\ref{fig:vg_amount}, we show the mean and standard deviation pointing game accuracy. The performance of the base vision-language objective does not exhibit a steady improvement with more text-image pairs. 
Although ReferIt performance increases, the performance on RefCOCO+ Test A remains mostly unchanged. 
Additionally, the performance on Flickr30k notably decreases, and there is a mixed effect on RefCOCO+ Test B, with half of the accuracy falling below the off-the-shelf ALBEF performance of 53.77\%.

In contrast, SelfEQ consistently leads to performance enhancements with more equivalent paraphrases.
Clear upward trends are observed on Flickr30k, ReferIt, and RefCOCO+ Test A as more data with corresponding equivalent paraphrases are added, meanwhile the gaps between the baselines generally widen. Notably, SelfEQ tuning maintains performance gains on Flickr30k, whereas the baselines performances drop. Although the trend on RefCOCO+ \mbox{Test B} is not consistently increasing, it is essential to emphasize that RefCOCO+ Test B is only a subset, and SelfEQ illustrates more stable and effective tuning performance on it, compared to the base vision-language objective.

These observations indicate that more equivalent paraphrases connecting with associated text phrases enable the model to acquire more valuable information during tuning. SelfEQ proves to be an effective and robust strategy for consistently improving performance with our generated paraphrases. With increased self-consistency augmented data, SelfEQ guides the model toward better grounding performance by enhancing its self-consistency capabilities.

Our method generates equivalent paraphrases for self-consistency enhancement, but it also contributes additional data for training.
To ascertain the specific impact of our SelfEQ tuning strategy, we run a control experiment on this variable.  As shown in Table~\ref{tab:extra_data}, we assess the model's performance when equivalent paraphrases are integrated as regular image-text pairs with vision-language objectives (Sec.~\ref{sec:loss}). This comparison reveals that merely augmenting the dataset with extra image-paraphrase pairs, without forming explicit linkages between the original text and its paraphrases, does not yield performance improvements.

\begin{table}[t!]
\centering
\small
\begin{tabular}{lccc}
\toprule
\textbf{Format} & \textbf{Objective} & \textbf{Flickr30k} & \textbf{ReferIt} \\
\midrule
- & $\mathcal{L}_{\mathrm{vl}}$ & 79.38 & 59.72 \\
\midrule
$C$ & $\mathcal{L}_{\mathrm{vl}}$ & 79.90 & 60.64 \\
$C$ & $\mathcal{L}_{\mathrm{SelfEQ}}$ & 81.28 &	62.04 \\
\midrule
$P$ & $\mathcal{L}_{\mathrm{vl}}$ & 81.18 & 61.18 \\
$P$ & $\mathcal{L}_{\mathrm{SelfEQ}}$ & \textbf{84.07} &	\textbf{62.75} \\
\bottomrule
\end{tabular}
\vspace{-0.05in}
\caption{Comparisons on data augmentation strategy for global-based captions in MS-COCO with or without the paraphrases. $C$ is the global-based captions from MS-COCO, and $P$ is our Vicuna-13B processed object-centric phrases separately. The first row is the off-the-shelf ALBEF performance before tuning.
\vspace{-0.2in}
}
\label{tab:chunk}
\end{table}

\vspace{0.05in}
\noindent
\textbf{Data Augmentation.}  
For global-based captions in MS-COCO, we preprocess the captions $C$ to object-centric short phrases $P$ via LLM-prompting. As shown in Table~\ref{tab:chunk}, tuning with phrases $P$ leads to better performance, benefiting both the vision-language objective ($\mathcal{L}_{\mathrm{vl}}$) and our self-consistency equivalence tuning objective ($\mathcal{L}_{\mathrm{SelfEQ}}$). This improvement is probably attributed to short phrases allowing the model to focus on a specific region rather than the entire scene, aligning more closely with the objective of visual grounding. By utilizing equivalent paraphrases with our SelfEQ objective (row 3 and 5), phrase chunking helps SelfEQ even more, indicating the important role of equivalent paraphrases in promoting self-consistency and grounding ability.

\vspace{0.05in}
\noindent
\textbf{Objective.} 
Table~\ref{tab:loss} evaluates each component within our self-consistency equivalence tuning objective. The \(\mathcal{L}_{\mathrm{sim}}\) loss targets pixel-wise similarity, ensuring that maps for a caption and its equivalent paraphrase are identical. However, focusing solely on pixel-level similarity may neglect the precise spatial positioning of objects. To address this, the \(\mathcal{L}_{\mathrm{cst}}\) loss is proposed to identify the most likely correct object positions within the two maps (\textit{i.e.}, RoI). It then encourages the model to yield consistently high attention scores within the RoI. By integrating both \(\mathcal{L}_{\mathrm{sim}}\) and \(\mathcal{L}_{\mathrm{cst}}\), self-consistency equivalence tuning objective fosters the model to not only align global similarities but also to pinpoint accurate object locations through the mutual supervision provided by a caption and its paraphrase, thereby enhancing self-consistency and accuracy.

\begin{table}[t!]
\centering
\small
\begin{tabular}{ccccccc}
\toprule
\multirow{2}{*}{\textbf{$\mathcal{L}_{\mathrm{sim}}$}} & \multirow{2}{*}{\textbf{$\mathcal{L}_{\mathrm{cst}}$}} & \multicolumn{2}{c}{\textbf{RefCOCO+}} & \multirow{2}{*}{\textbf{Flickr30k}} & \multirow{2}{*}{\textbf{ReferIt}} \\
\cmidrule(lr){3-4}
 & & \textbf{Test A} & \textbf{Test B} \\
\midrule
\checkmark & & 66.42 & 47.21 & 68.26 & 55.96 \\
& \checkmark & 73.33 & 55.88 & 80.94 & 66.57 \\
\checkmark & \checkmark & \textbf{75.10} & \textbf{55.49} & \textbf{81.90} & \textbf{67.40} \\
\bottomrule
\end{tabular}
\vspace{-0.05in}
\caption{Ablation studies on objective components of self-consistency equivalence tuning objective $\mathcal{L}_{\mathrm{SelfEQ}}$.}
\label{tab:loss}
\vspace{-0.1in}
\end{table}

\section{Conclusion}
In this paper, we propose a novel weakly-supervised tuning approach coupled with a data augmentation strategy to enhance the localization capabilities of a purely image-text pair supervised model through self-consistency.
Using an open-source LLM, we expand a dataset with equivalent paraphrases tailored to be object-centric. The augmented data is used to finetune a base model employing our novel self-consistency equivalence tuning objective. Our approach has been rigorously validated across pretraining on diverse datasets, ranging from region-based captions (\textit{i.e.}, VG) to global-based captions (\textit{i.e.}, COCO). 
Our method achieves superior and self-consistent performance on three benchmarks and is even competitive with some box-supervised methods. 

\clearpage
{
    \small
    \bibliographystyle{ieeenat_fullname}
    \bibliography{main}
}

\clearpage
\appendix

\section*{Appendix}
We present details about our two-level LLM prompts to obtain high-quality paraphrases for region-centric phrases, provide examples of such extracted paraphrases, conduct additional evaluations on in-the-wild captions on CC3M, justify our selection of ALBEF as our base model, evaluate in more detail the choice of object-centric captions, and provide additional qualitative results.

\section{LLM-Prompting Details}
\begin{figure}[htbp] 
    \centering
    \includegraphics[width=\linewidth]{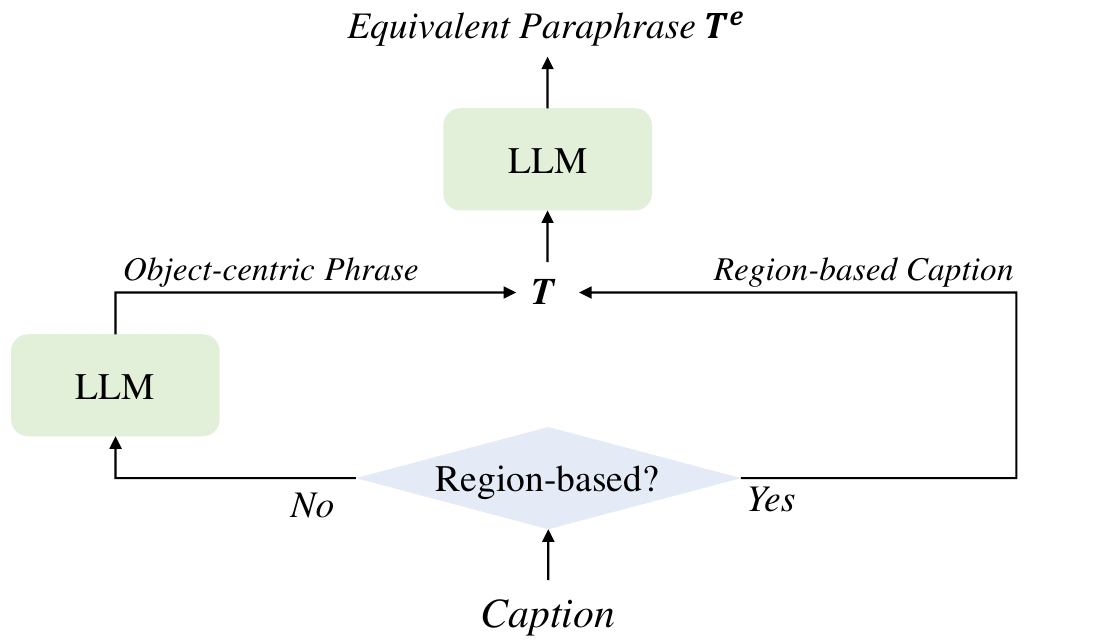}
    \vspace{-0.2in}
    \caption{An overview of our two-level self-consistency data augmentation approach. For global-based captions, we use an LLM to chunk object-centric phrases $T$ with our first-level prompting. The obtained phrases or region-based captions $T$ are further input to an LLM in the second-level prompting to obtain equivalent paraphrases $T^e$.}
    \vspace{-0.1in}
    \label{fig:prompt_pipeline}
\end{figure}

This section presents prompting details and generated examples of our proposed two-level self-consistency data augmentation method (refer to~\crefordefault{sec:data}{Section~\blue{3.3} in the main text}). As shown in Figure~\ref{fig:prompt_pipeline}, we classify textual annotations from existing datasets into global-based captions that describe the entire image, and \mbox{region-based} captions that describe a specific region within the image. In our experiments, we identify captions from Visual Genome (VG)~\cite{krishna2017visual} as \mbox{region-based} captions, while captions from MS-COCO~\cite{lin2014microsoft} and Conceptual Captions 3M (CC3M)~\cite{sharma2018conceptual} are referred to as \mbox{global-based} captions. 

\subsection{Prompt Design}
We adopt an in-context learning~\cite{brown2020language} strategy in our LLM-prompt design. Each pair of our prompts encompasses a query text \textbf{Q} and an expected answer \textbf{A}. 
The query texts \textbf{Q} were selected and modified based on generation quality and successful rate on a small validation subset.

\begin{figure}[t] 
    \centering
    \includegraphics[width=\linewidth]{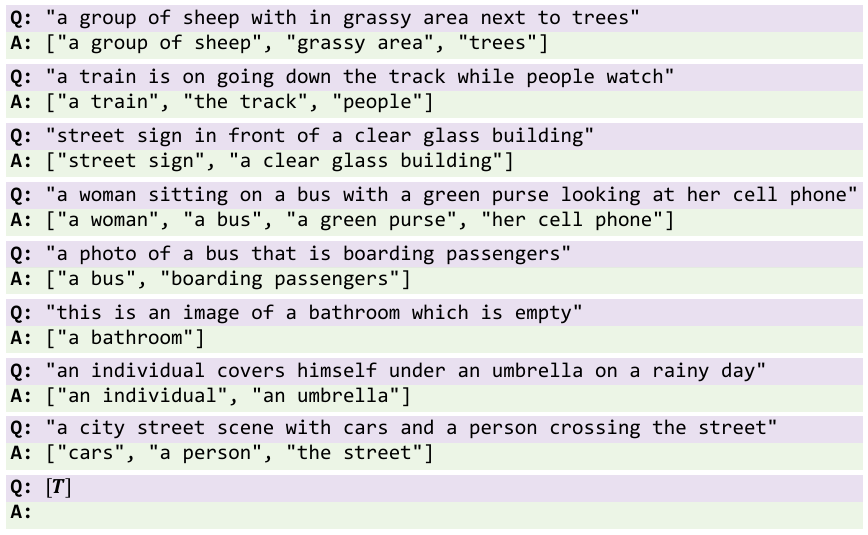}
    \vspace{-0.2in}
    \caption{In-context few-shot LLM-Prompt for our first-level self-consistency data augmentation. We leverage an LLM for phrase chunking to obtain object-centric captions from captions that describe images globally. 
    $[T]$ is the query text. \textbf{A} in the last row is followed by the output.}
    \vspace{-0.2in}
    \label{fig:chunk_prompt}
\end{figure}

\vspace{-0.1in}
\paragraph{Phrase Chunking.}
To obtain object-centric phrases, we prompt an LLM for global-based captions as shown in Figure~\ref{fig:chunk_prompt}. Compared to previous phrase chunking methods~\cite{zhai2017neural}, this LLM-prompt-based approach aligns more closely with our objective. In conventional phrase chunking, the output chunks include nouns, verbs, and prepositional phrases. However, our SelfEQ method focuses on object-centric phrases rather than verbs or prepositions. Additionally, there are some abstract nouns such as ``photo", ``image" and ``scene" that are commonly used in annotations but do not benefit visual grounding. We select several captions containing abstract nouns, verbs, and prepositions as our query texts in our prompts to guide the LLM to generate the chunks in a task-specific way.

\begin{figure}[t!] 
    \centering
    \includegraphics[width=\linewidth]{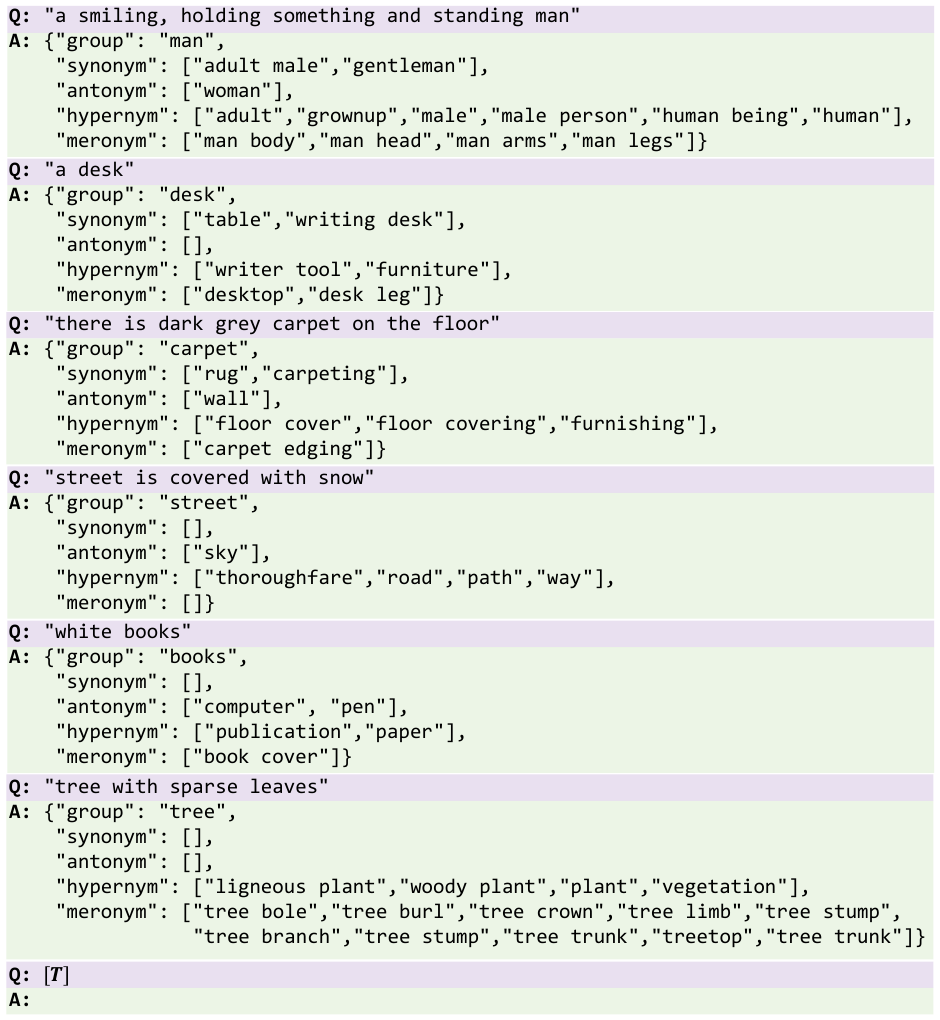}
    \vspace{-0.2in}
    \caption{In-context few-shot LLM-Prompt for our second-level self-consistency data augmentation on Visual Genome (VG). We leverage an LLM to generate paraphrases for the given textual descriptions (region-based captions). $[T]$ is the regional-based caption, and \textbf{A} in the last row is followed by the expected output.}
    \label{fig:vg_paraphrase_prompt}
    \vspace{-0.18in}
\end{figure}

\vspace{-0.1in}
\paragraph{Paraphrase Generation.} Our second-level prompts aim to generate paraphrased captions through substitute words and for this purpose we use an in-context few-shot prompt with an LLM.
Figure~\ref{fig:vg_paraphrase_prompt} shows the specific prompt we use for input region-based captions.  
As in-context examples, we select four captions with different sentence structures or where the primary object is playing different syntactic roles: A complex noun phrase (set 1), an existential structure (set 3), a subject followed by a passive verb phrase (set 4), and a noun phrase modified by a prepositional phrase (set 6). We also add two short noun phrases (sets 2, 5) to represent relevant captions. For the expected output \textbf{A}, we first detect the primary object (\ie, the ``group" field) in the query text, then use WordNet synsets~\cite{miller1995wordnet} to generate relationships (\ie, the ``synonym", ``antonym", ``hypernym", and ``meronym" fields) for the primary object. We further remove inaccurate words as the final expected output \textbf{A} in our LLM-prompt.

\begin{figure}[t!] 
    \centering
    \includegraphics[width=\linewidth]{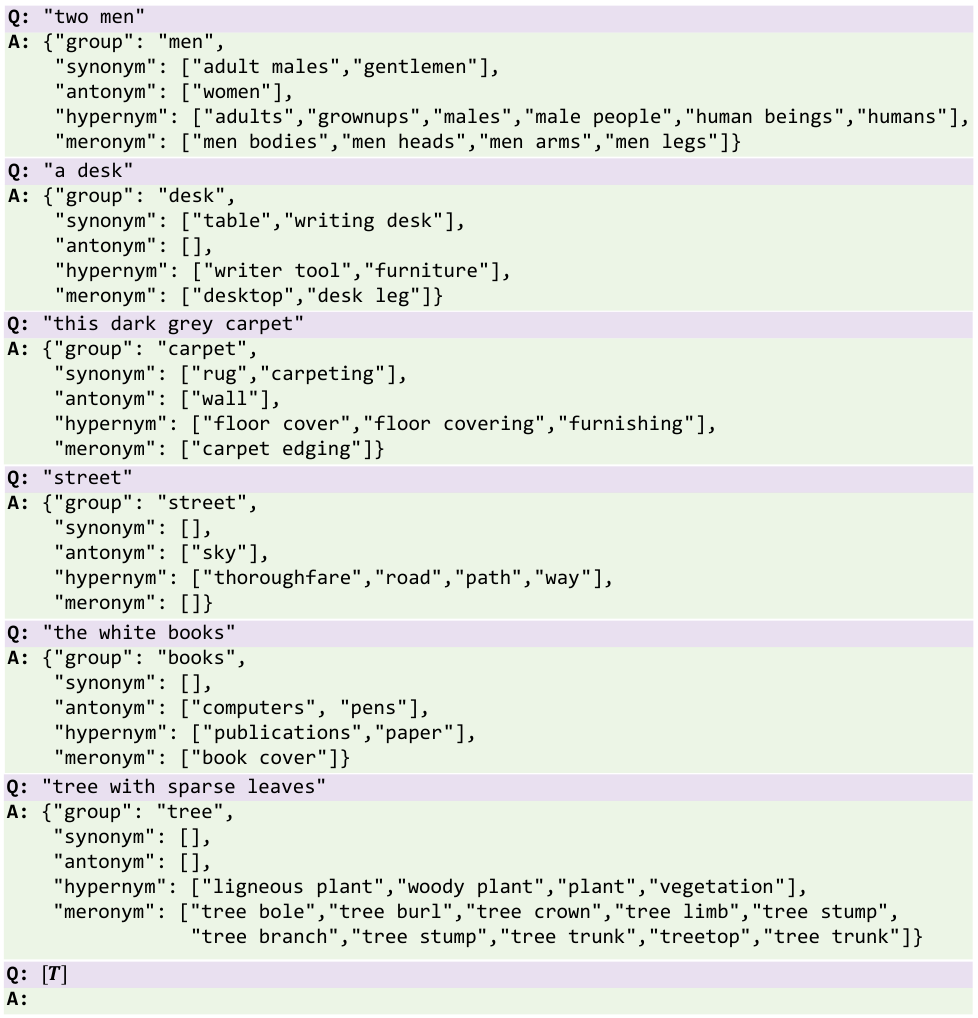}
    \vspace{-0.2in}
    \caption{In-context few-shot LLM-Prompt for our second-level self-consistency data augmentation on MS-COCO. We leverage an LLM to generate paraphrases for the given textual descriptions (object-centric phrases). $[T]$ is the object-centric phrase obtained from the first-level phrase chunking. \textbf{A} in the last row is followed by the expected output.}
    \label{fig:coco_paraphrase_prompt}
    \vspace{-0.18in}
\end{figure}

Figure~\ref{fig:coco_paraphrase_prompt} shows our in-context few-shot LLM-prompt for object-centric phrases obtained from our first-level data augmentation. It is based on our LLM-prompt for region-based captions (Figure~\ref{fig:vg_paraphrase_prompt}), but we further chunk the query texts into object-centric phrases for adaption and modify them (\eg, articles, pronouns) empirically on a small validation subset.

\subsection{Generated Data Examples}
This section presents generated examples from our two-level self-consistency augmentation approach. We apply paraphrase generation for VG and both phrase chunking and paraphrase generation for MS-COCO and CC3M.

\begin{figure*}[htbp] 
\vspace{-0.25in}
    \centering
    \includegraphics[width=0.45\textwidth]{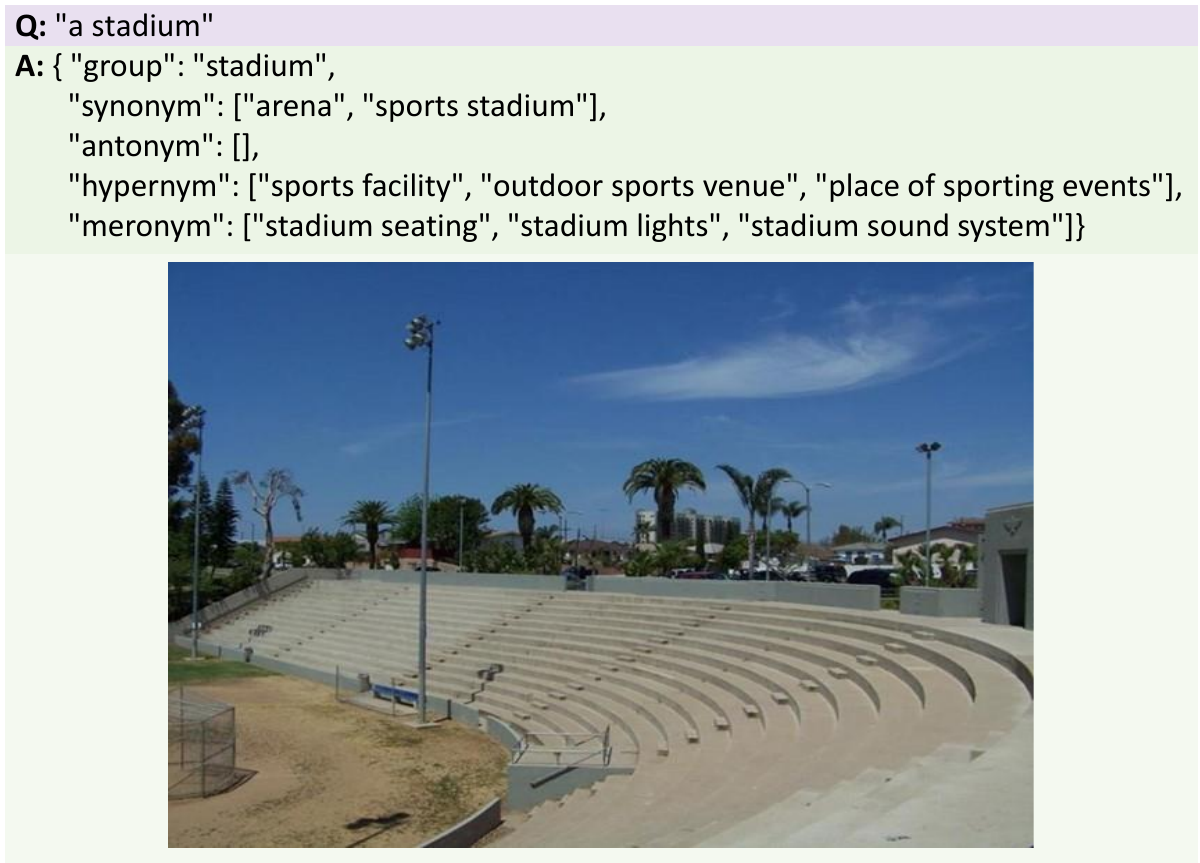} 
    \includegraphics[width=0.45\textwidth]{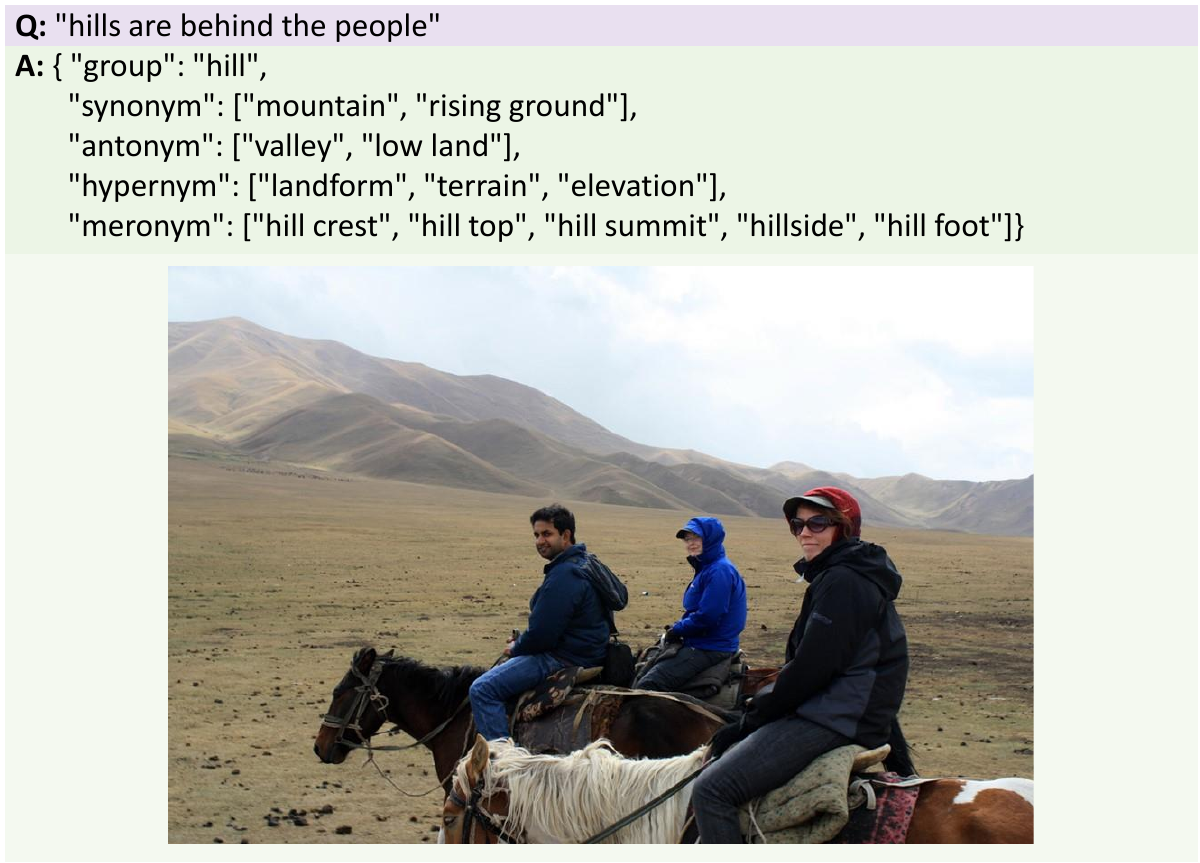}
    \includegraphics[width=0.45\textwidth]{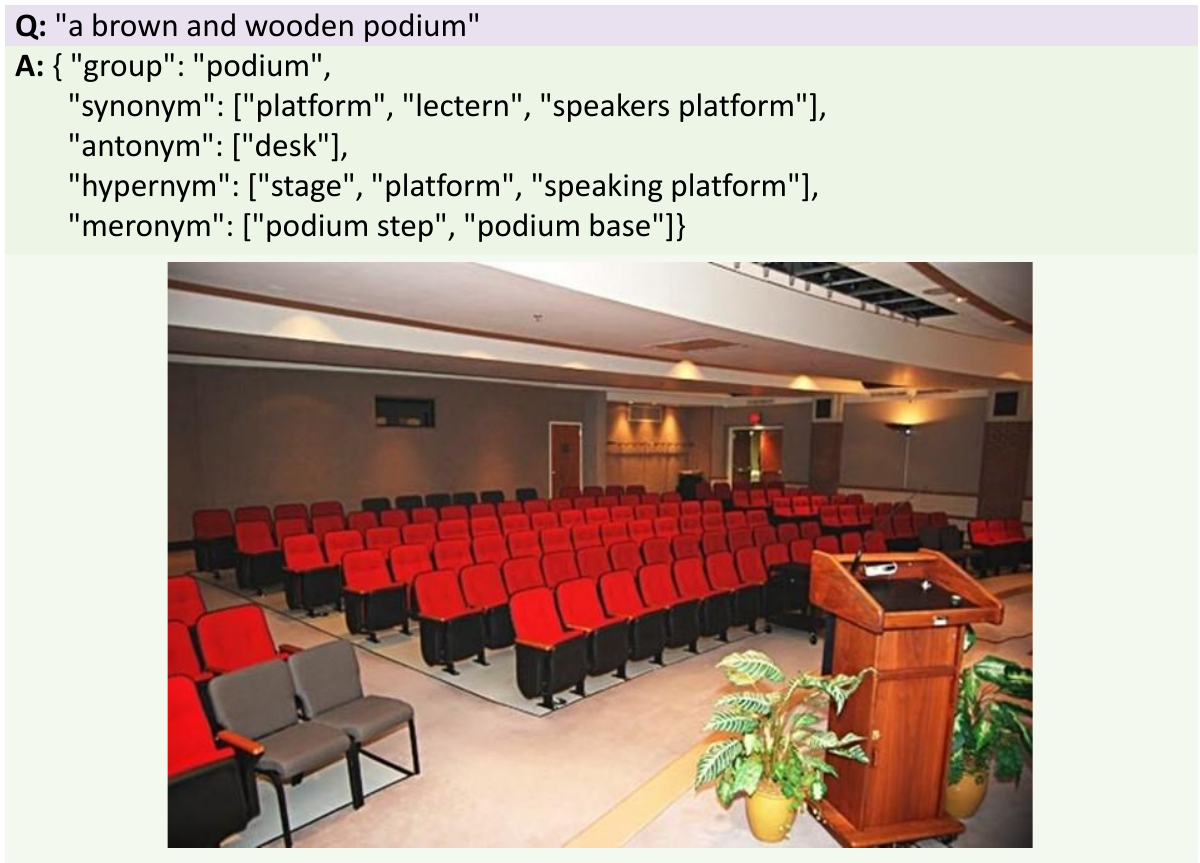} 
    \includegraphics[width=0.45\textwidth]{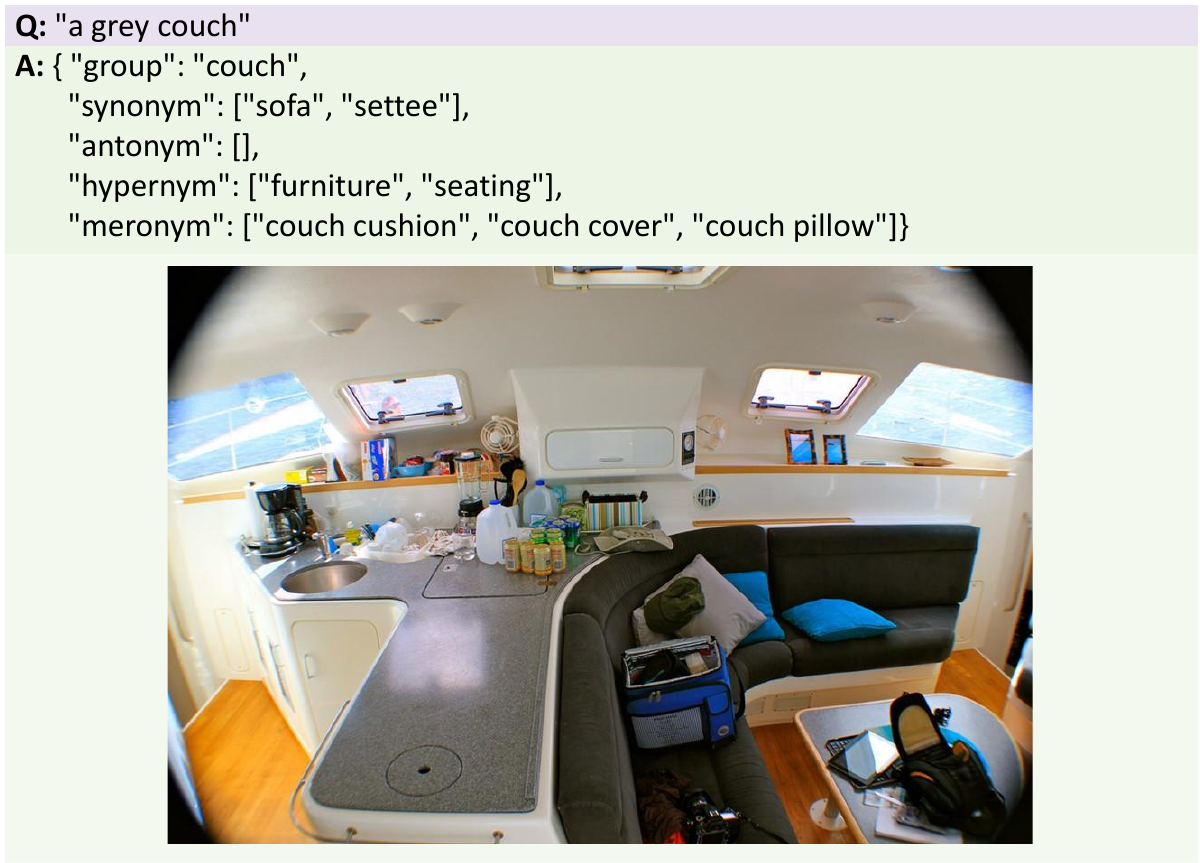}
    \includegraphics[width=0.45\textwidth]{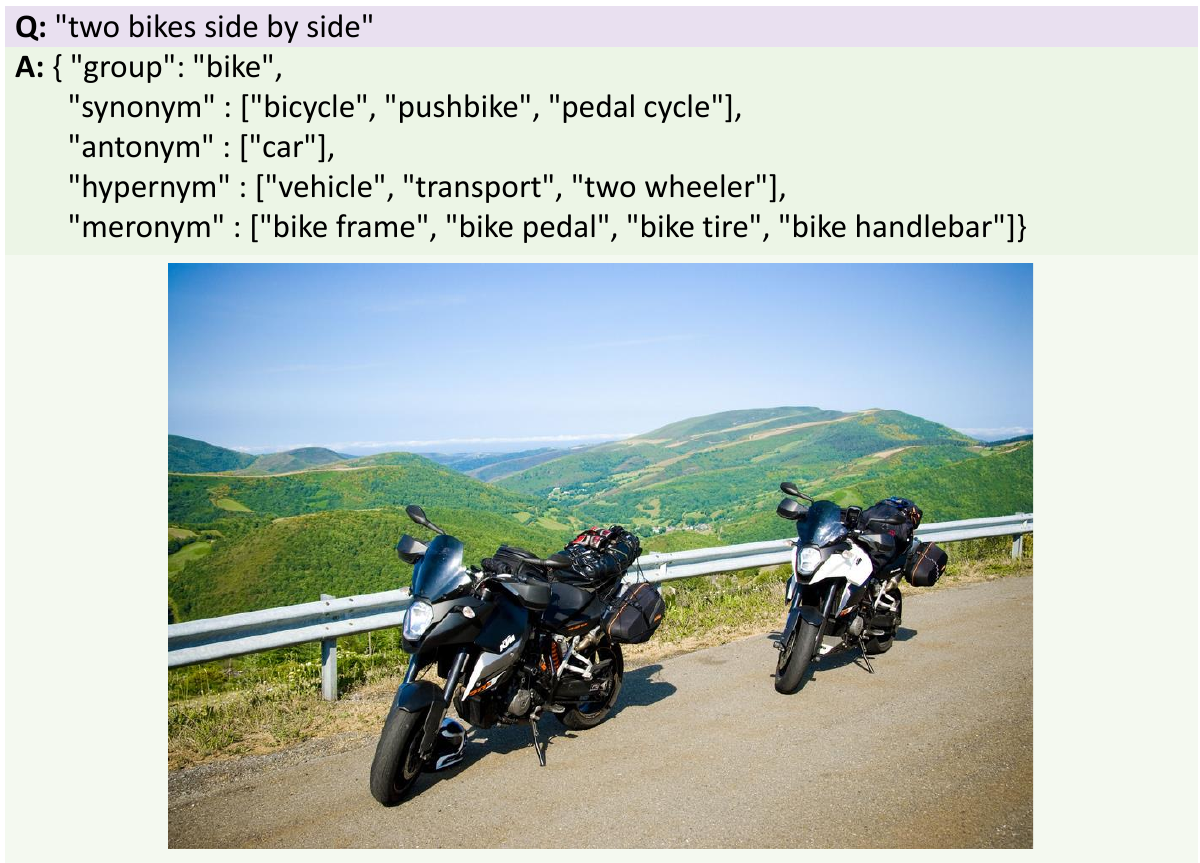} 
    \includegraphics[width=0.45\textwidth]{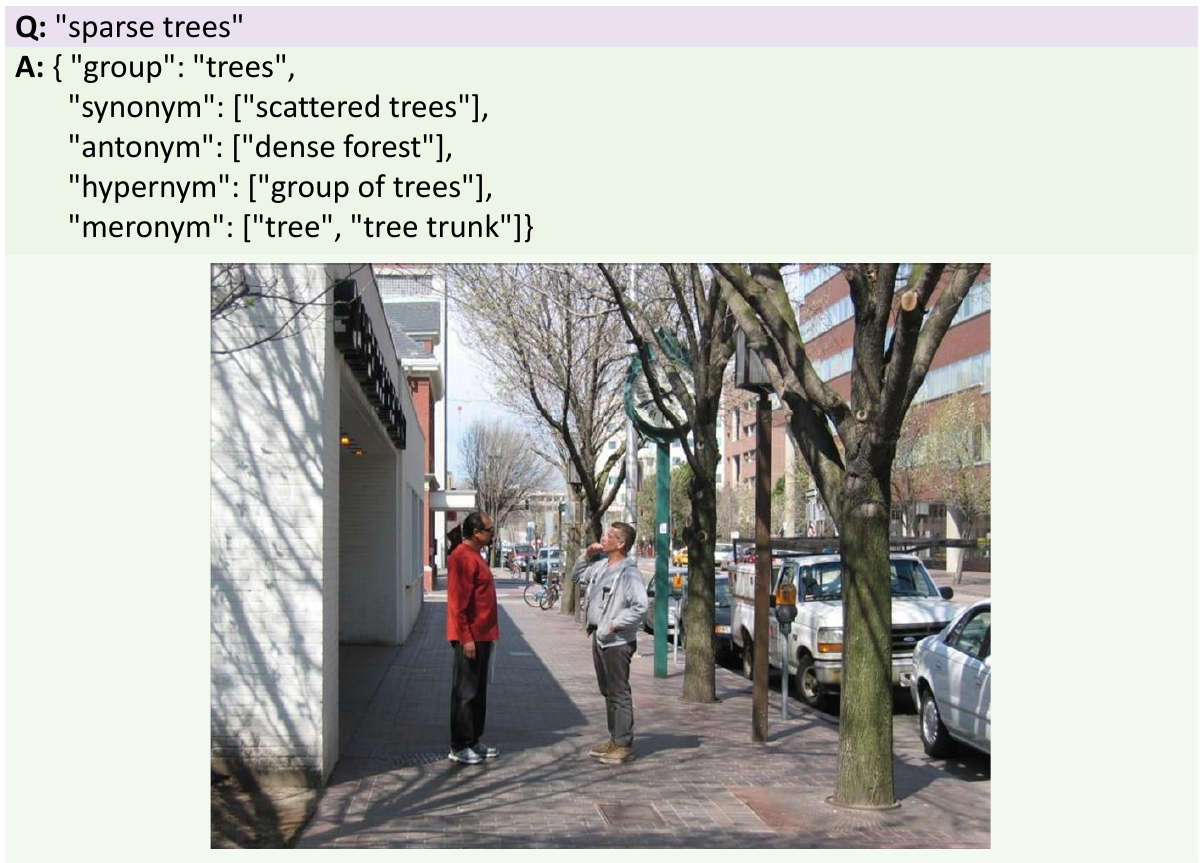} 
    \includegraphics[width=0.45\textwidth]{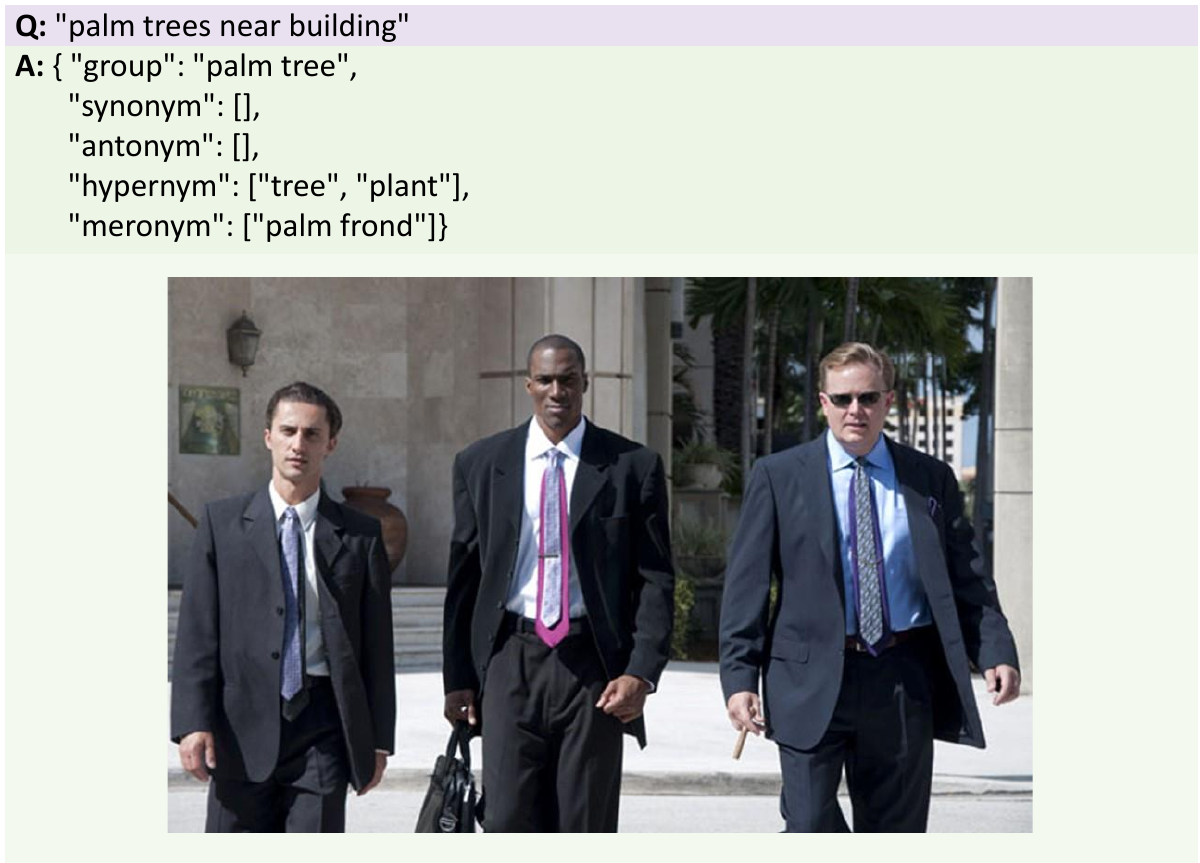} 
    \includegraphics[width=0.45\textwidth]{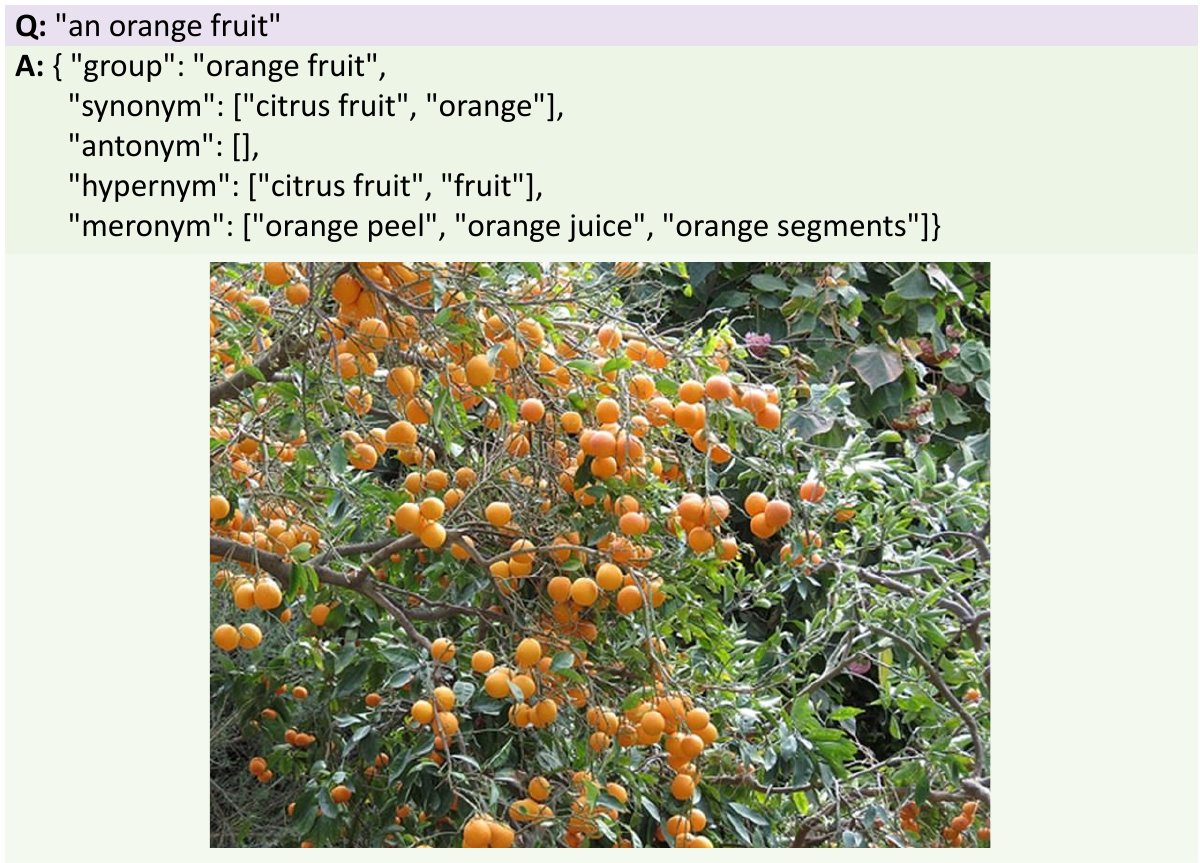}
    \vspace{-0.1in}
    \caption{LLM generated examples for VG. \textbf{Q} represents the query text associated with the image. \textbf{A} corresponds to the output of our second-level self-consistency data augmentation. ``group" denotes the detected primary object, and it further generates relationships such as ``synonym," ``antonym," ``meronym," and ``hypernym" regarding with the identified ``group."}
    \label{fig:vg_paraphrase_example_w_images}
\end{figure*}

\vspace{-0.1in}
\paragraph{Visual Genome.}
Figure~\ref{fig:vg_paraphrase_example_w_images} shows generated example paraphrases for VG. Since our LLM-prompt contains various sentence structures in the provided in-context examples, the generated data showcases the successful detection of primary objects (``group") for a variety of input captions. To ensure the quality of equivalent paraphrases, we allow the LLM to leave blanks in relevant fields if no appropriate words are available. In total, the LLM generates equivalent paraphrases for 74.56\% of captions for VG. The different types of equivalent paraphrases include general synonyms (\eg, ``bicycle", ``sofa"), formal or technical terms (\eg, ``pedal cycle"), colloquial or regional variants (\eg, ``pushbike", ``settee") and descriptive synonyms (\eg, ``scattered trees"). 

\begin{figure*}[htbp] 
\vspace{-0.25in}
    \centering
    \includegraphics[width=0.43\textwidth]{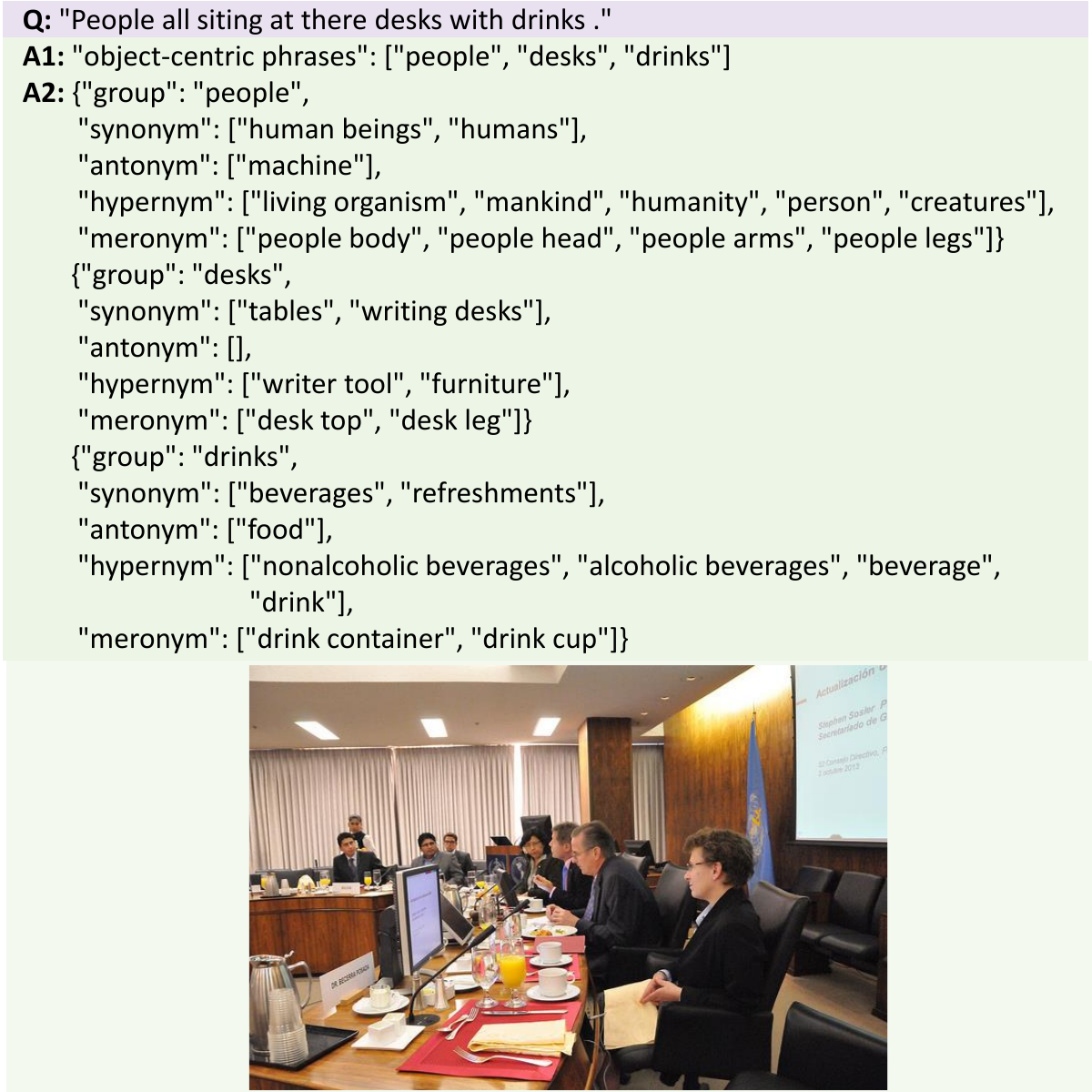} 
    \includegraphics[width=0.43\textwidth]{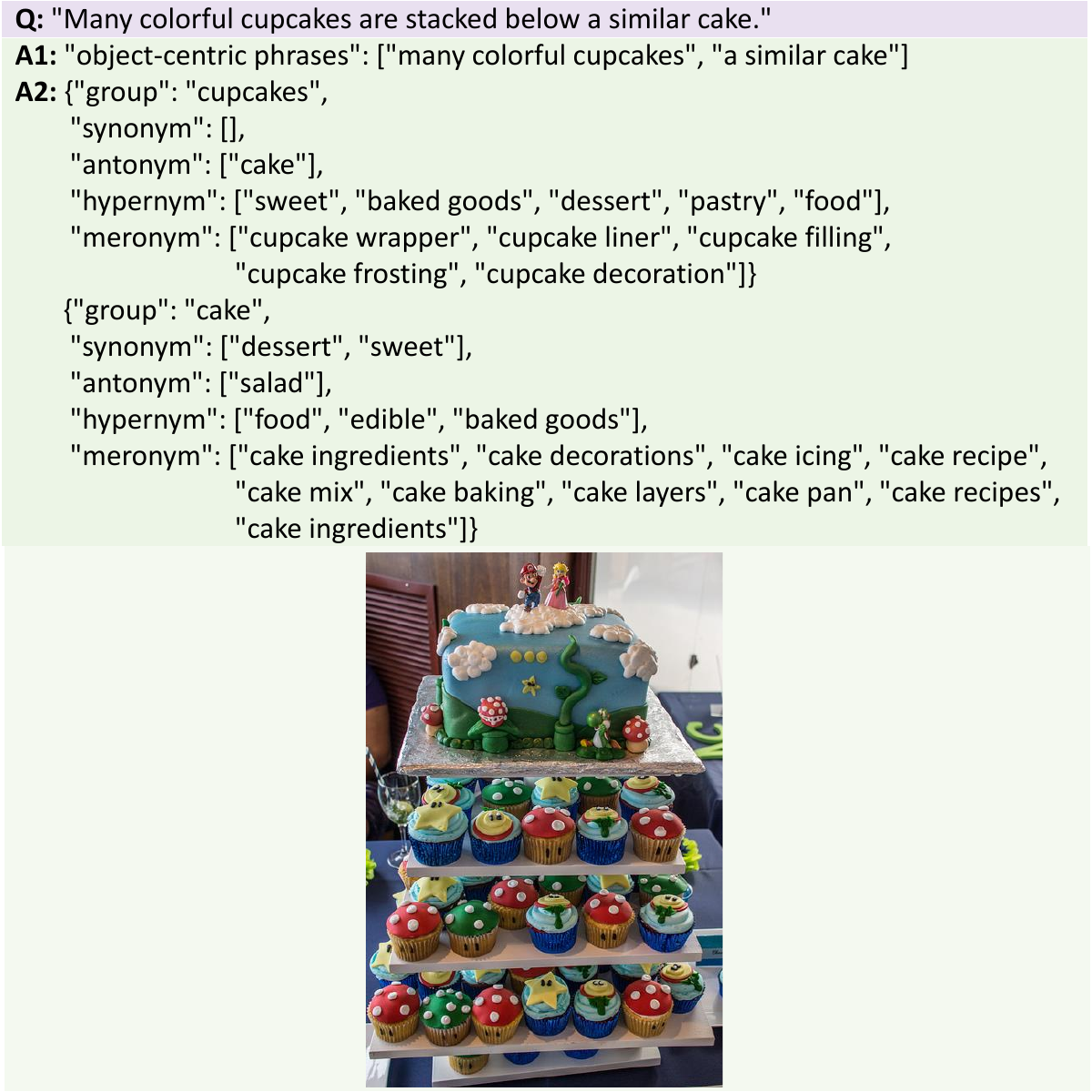}
    \includegraphics[width=0.43\textwidth]{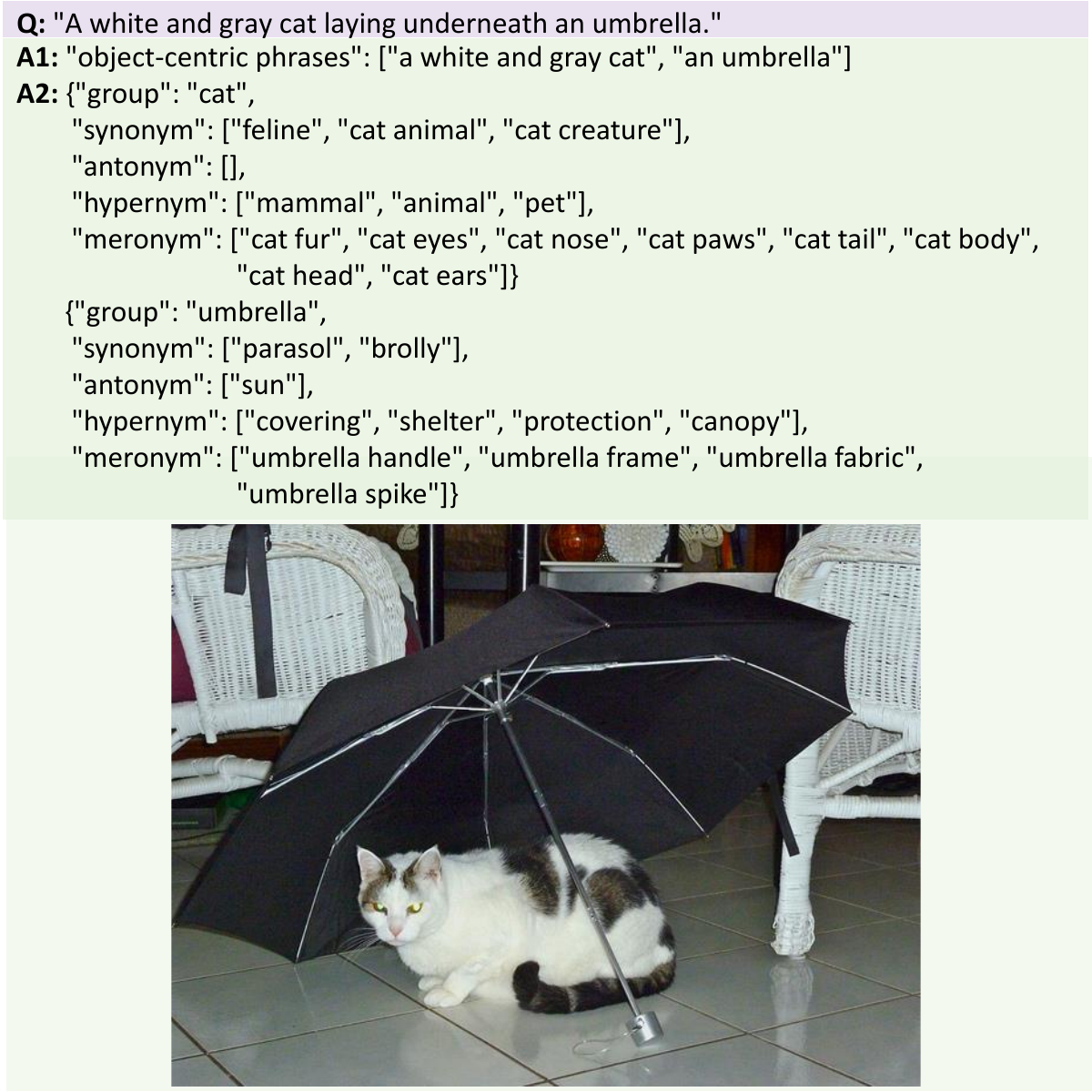} 
    \includegraphics[width=0.43\textwidth]{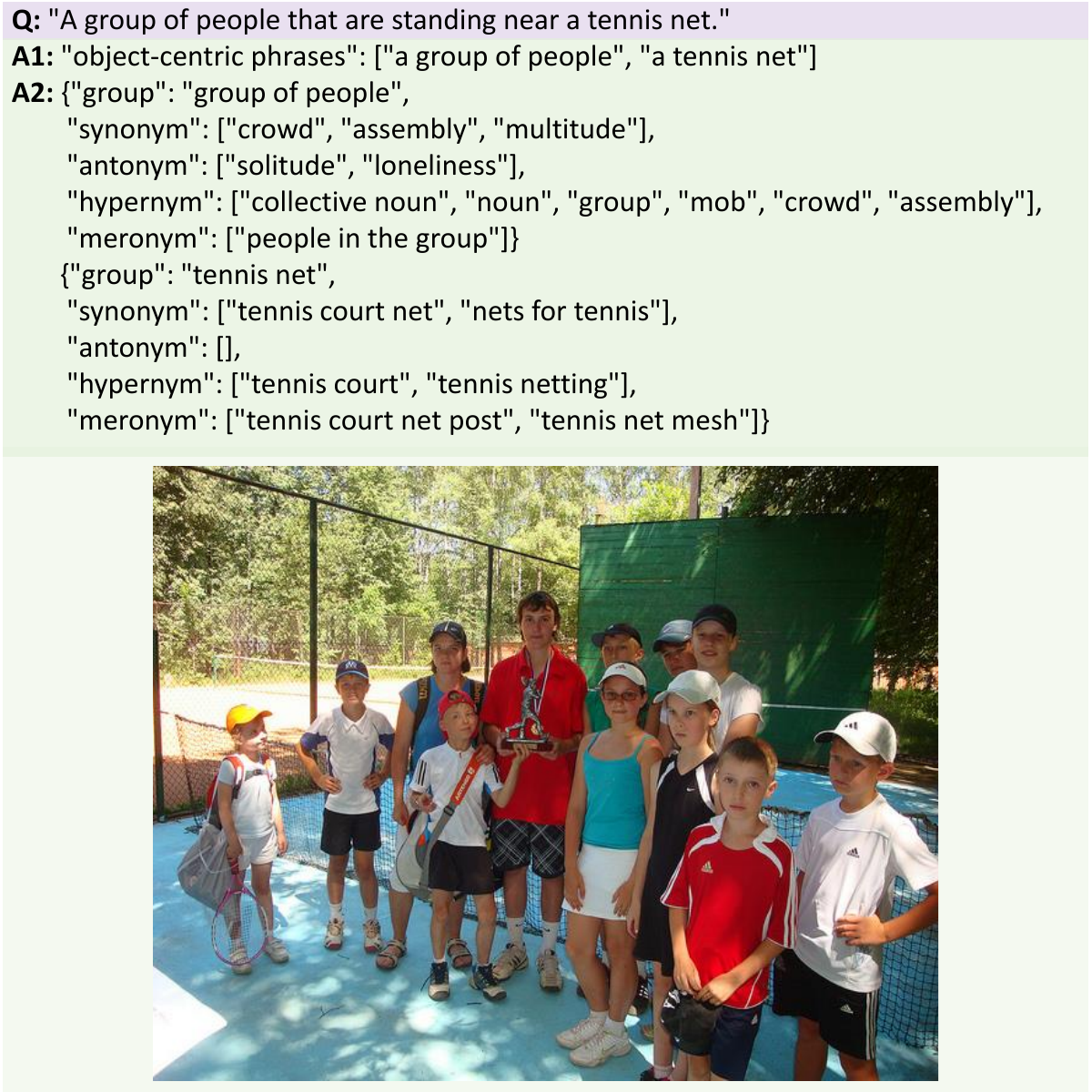}
    \includegraphics[width=0.43\textwidth]{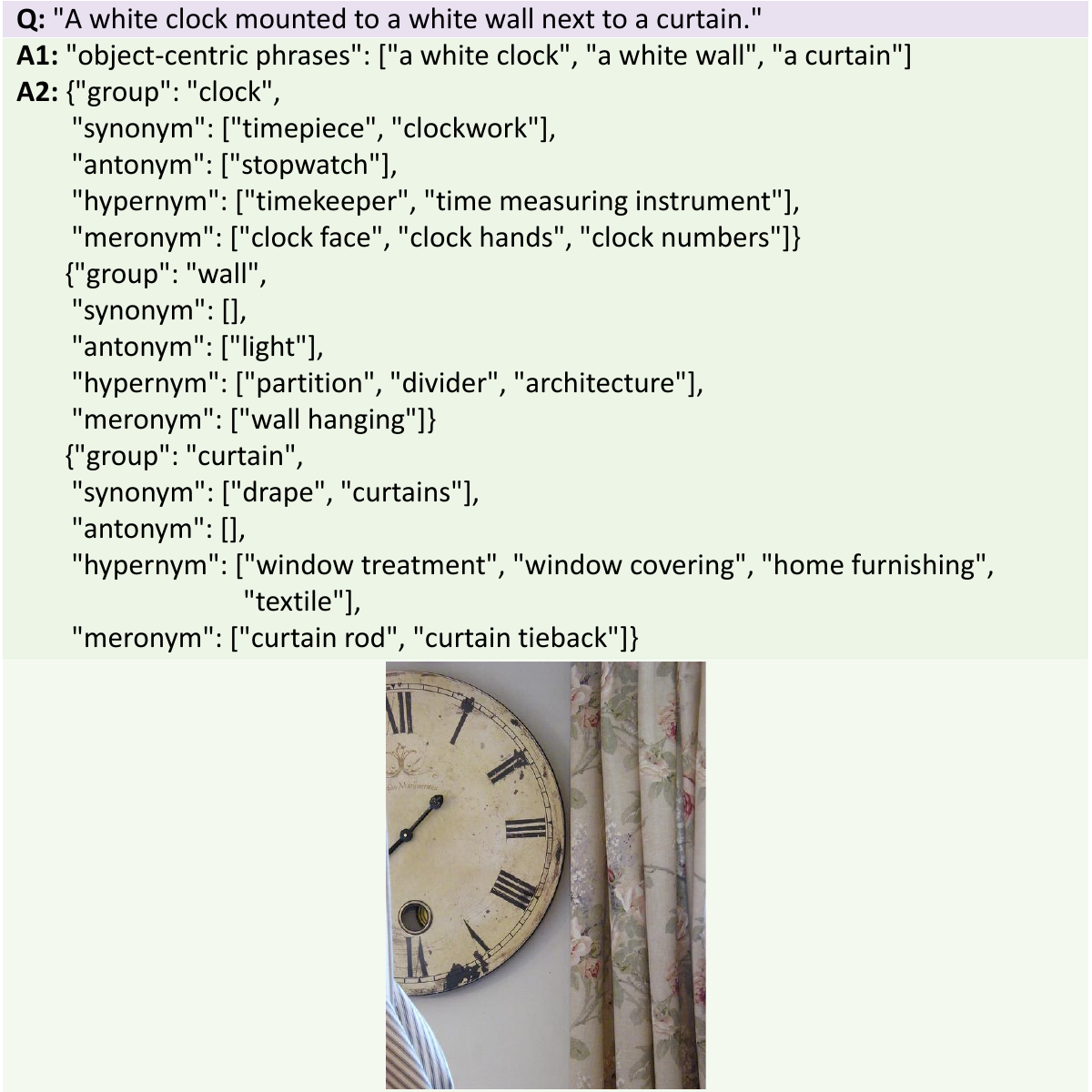} 
    \includegraphics[width=0.43\textwidth]{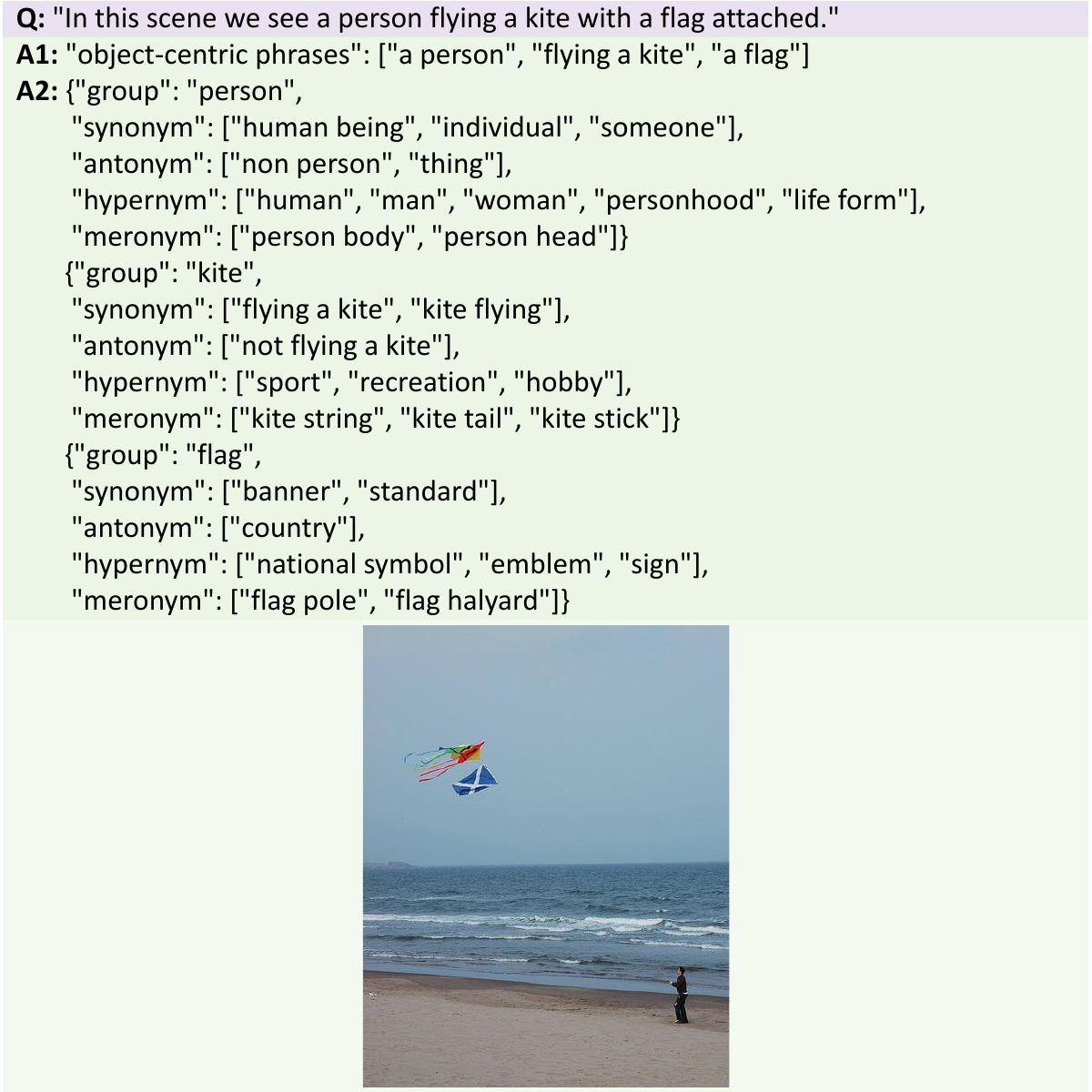}
    \vspace{-0.1in}
    \caption{LLM generated examples for MS-COCO. \textbf{Q} represents the query text associated with the image. \textbf{A1} is the object-centric phrase obtained from the first-level self-consistency data augmentation, while \textbf{A2} corresponds to the second level. For each object-centric phrase in \textbf{A1}, LLM detects primary objects ``group" and generates relevant relationships in \textbf{A2}.}
    \label{fig:coco_paraphrase_example_w_images}
\end{figure*}

\begin{figure*}[tbp] 
\vspace{-0.25in}
    \centering
    \includegraphics[width=0.43\textwidth]{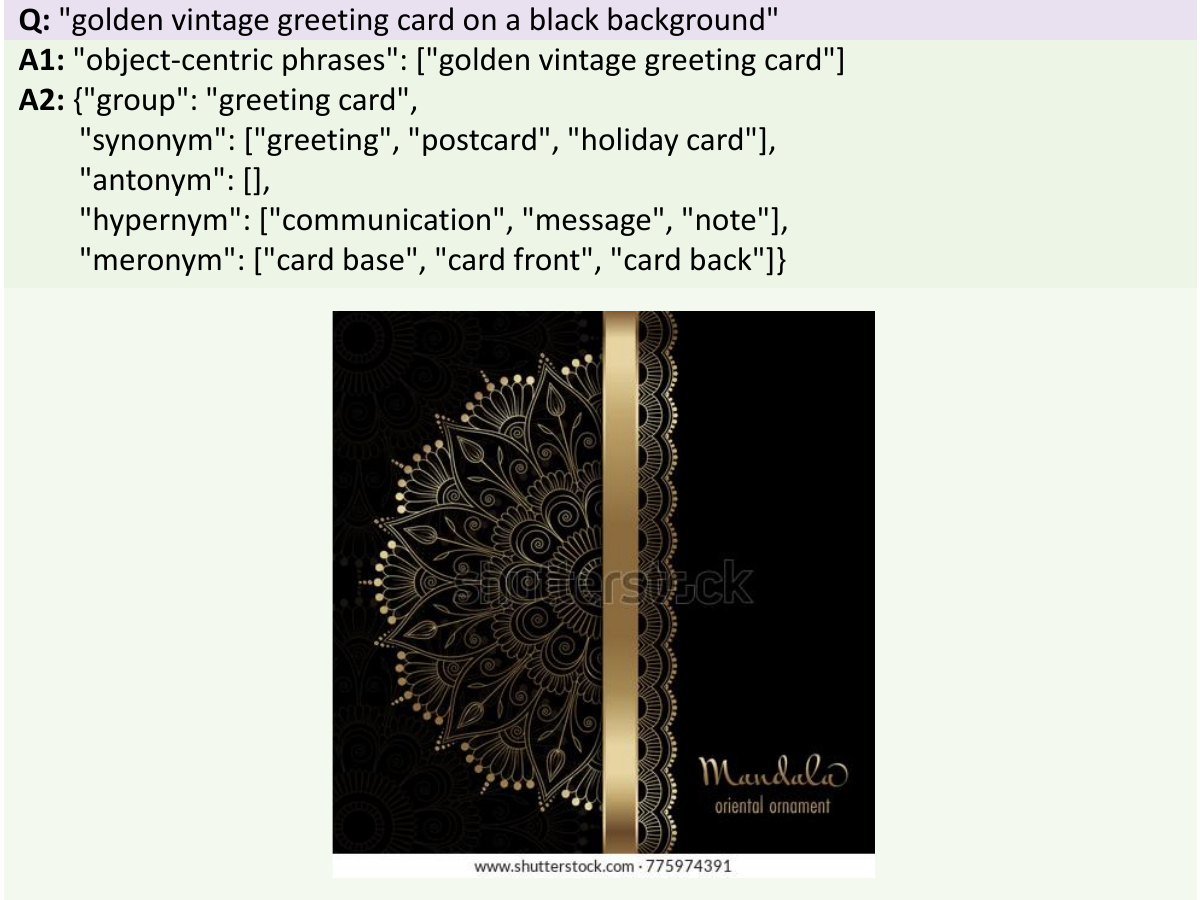} 
    \includegraphics[width=0.43\textwidth]{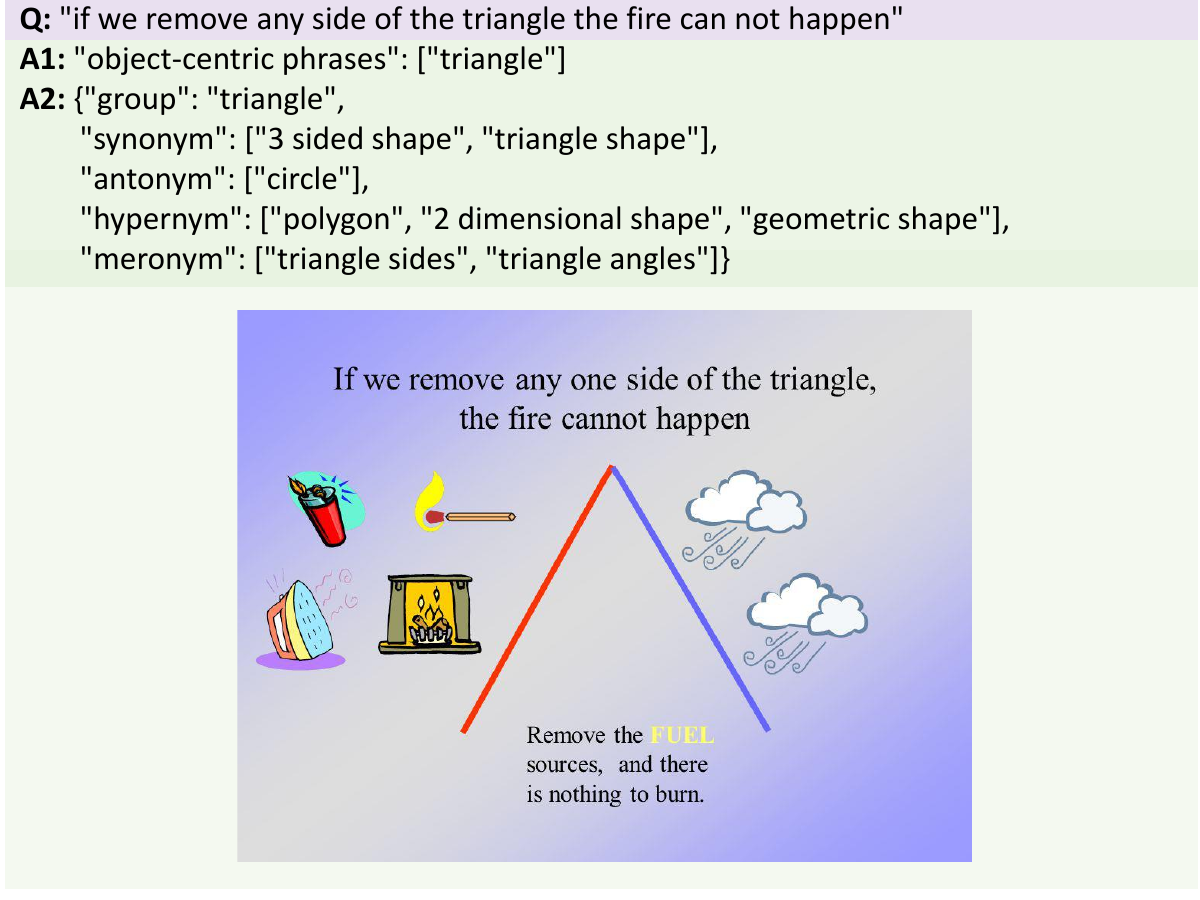}
    \includegraphics[width=0.43\textwidth]{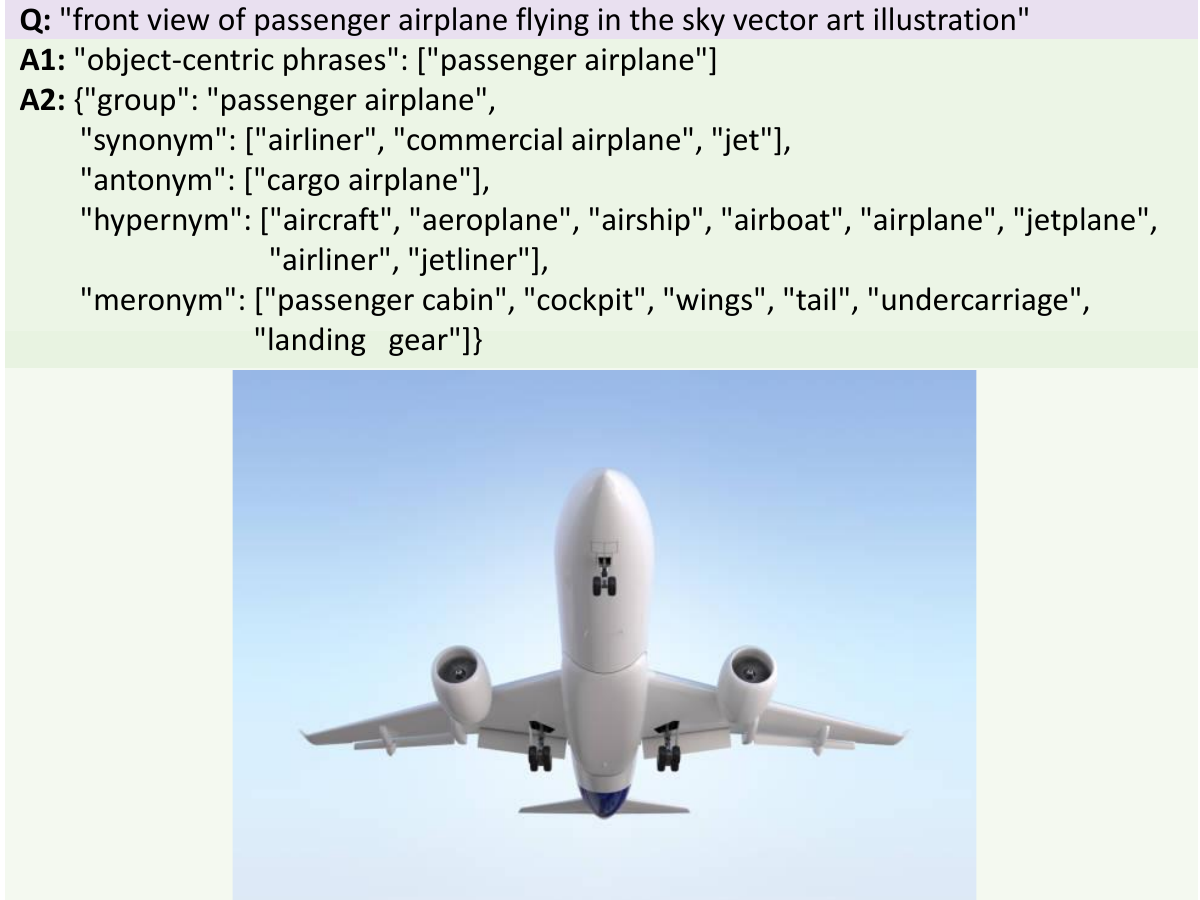} 
    \includegraphics[width=0.43\textwidth]{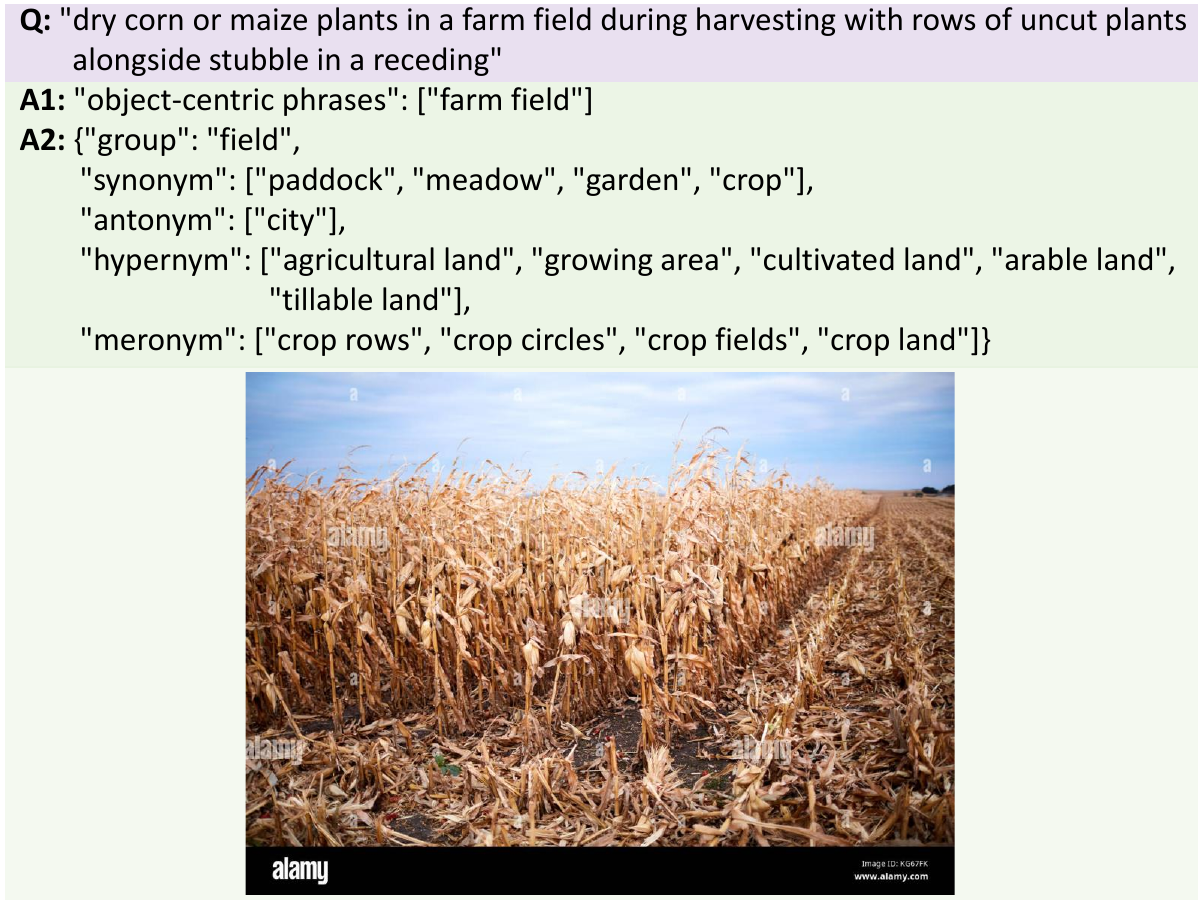}
    \includegraphics[width=0.43\textwidth]{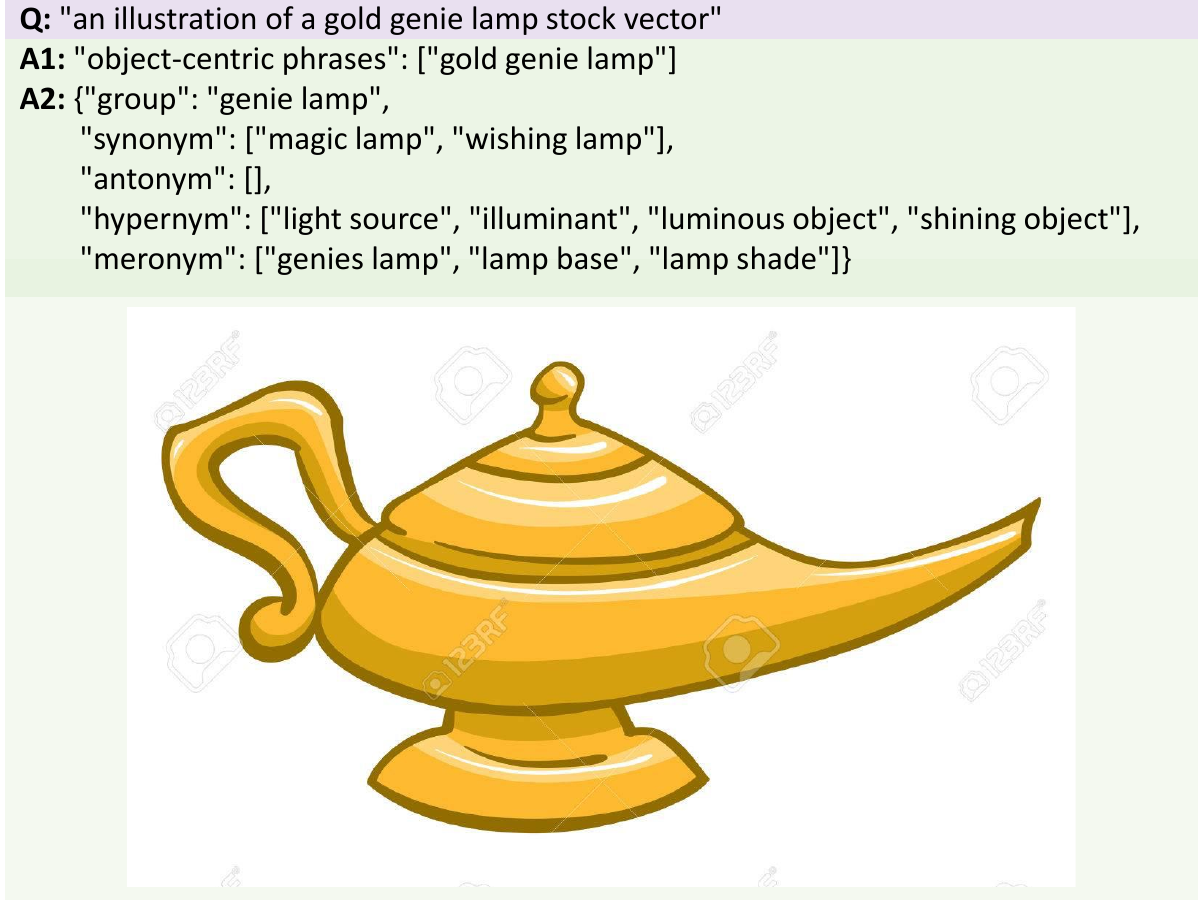} 
    \includegraphics[width=0.43\textwidth]{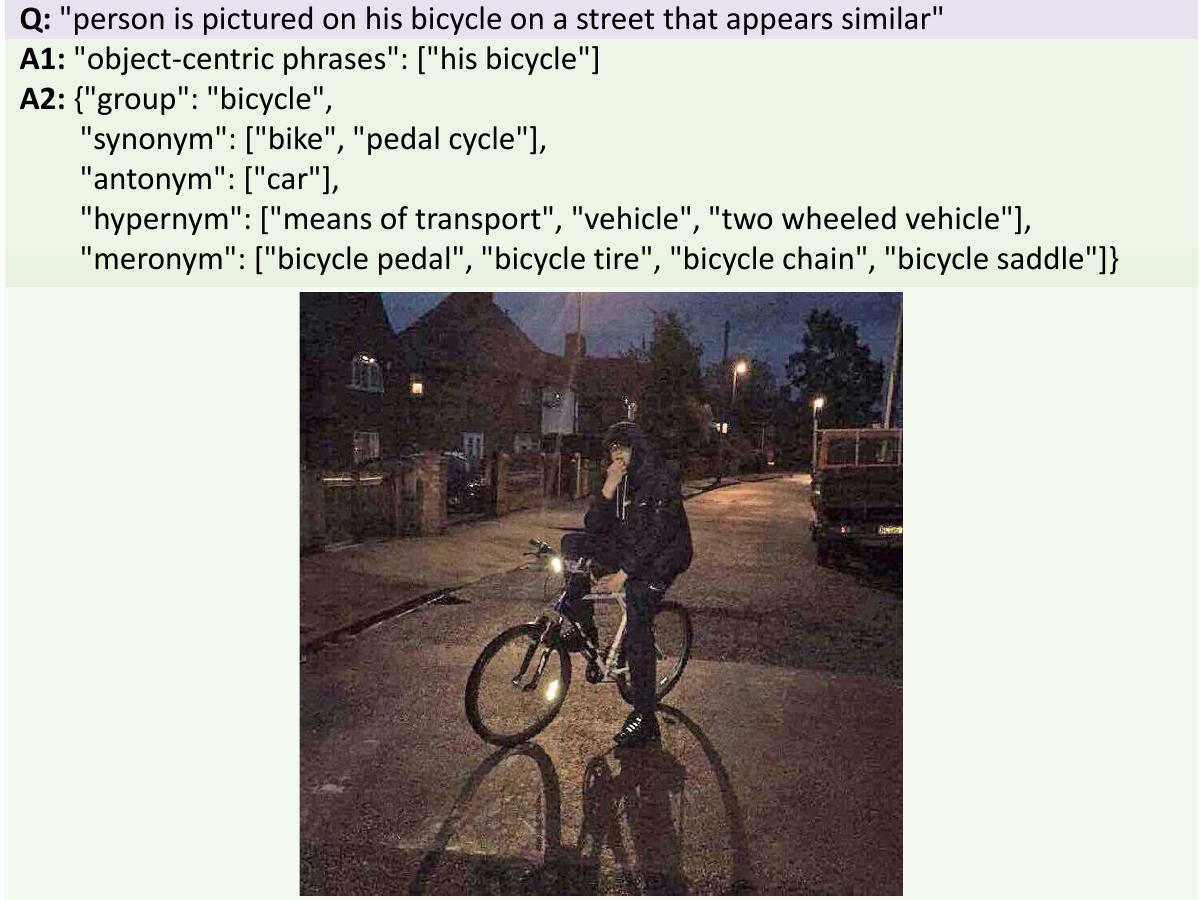} 
    \includegraphics[width=0.43\textwidth]{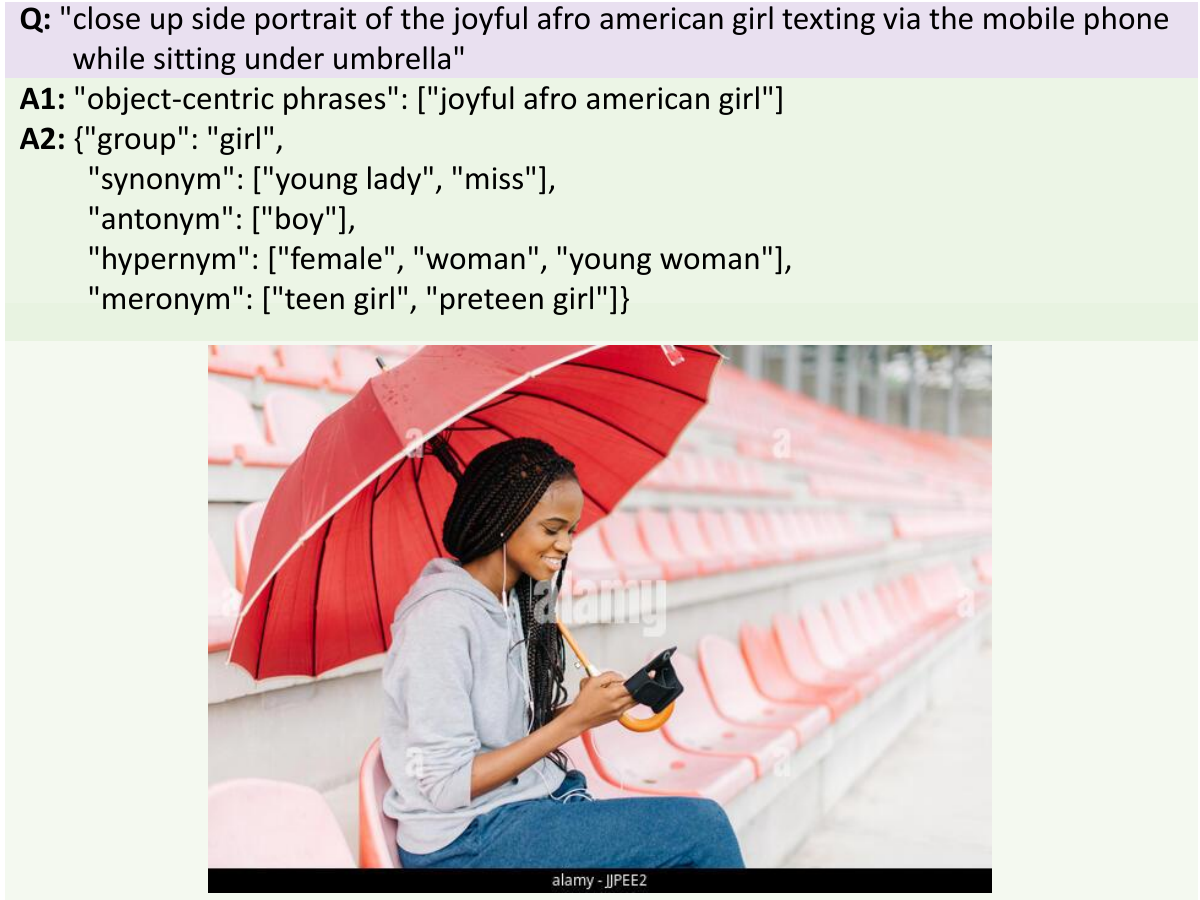} 
    \includegraphics[width=0.43\textwidth]{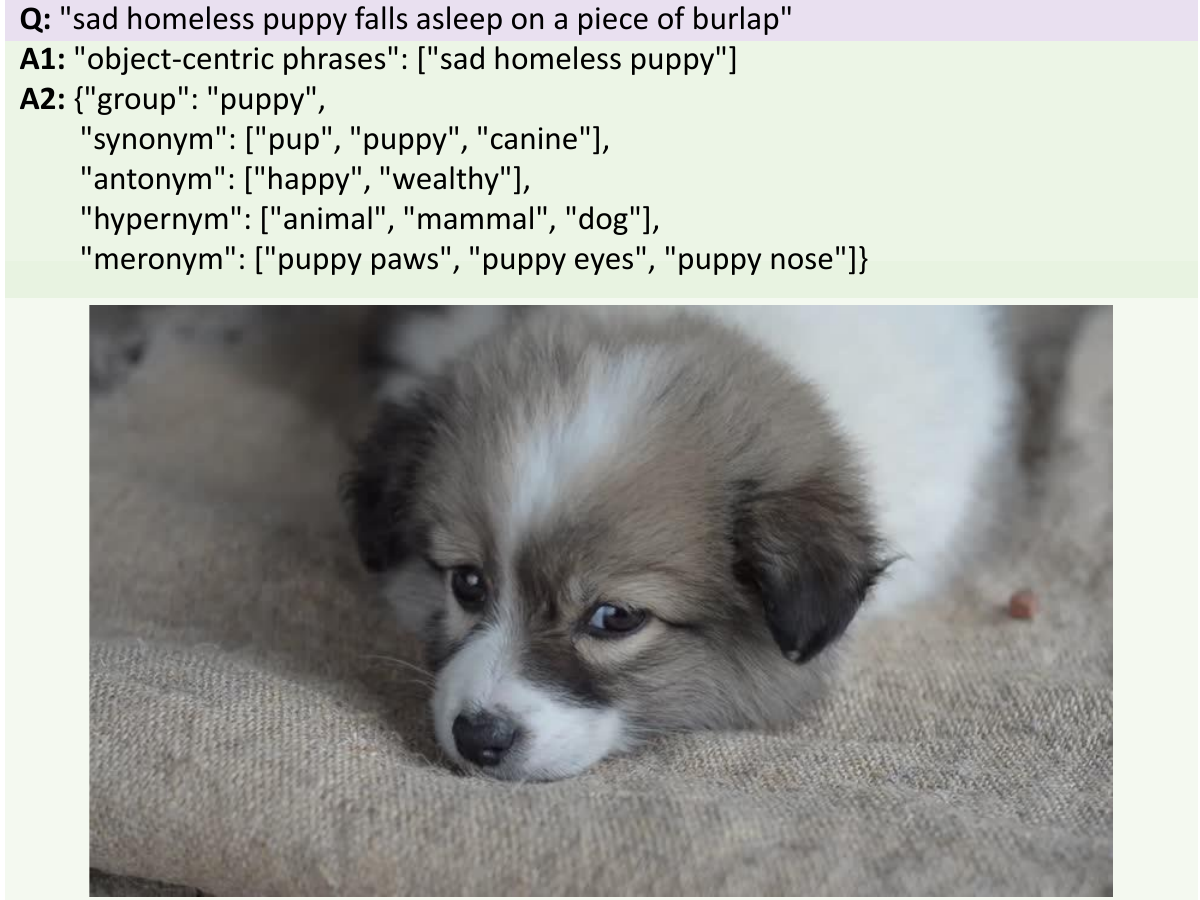}
    \vspace{-0.1in}
    \caption{LLM generated examples for CC3M. \textbf{Q} represents the query text associated with the image. \textbf{A1} is the object-centric phrase obtained from the first-level self-consistency data augmentation, while \textbf{A2} corresponds to the second level. For each object-centric phrase in \textbf{A1}, LLM detects primary objects ``group" and generates relevant relationships in \textbf{A2}.}
    \label{fig:cc3m_paraphrase_example_w_images}
\end{figure*}

\vspace{-0.15in}
\paragraph{MS-COCO.}
Generated examples for MS-COCO are shown in Figure~\ref{fig:coco_paraphrase_example_w_images}. Captions in MS-COCO describe the entire images, so we augment them with our two-level phrase augmentation: Phrase chunking and equivalent paraphrase generation. In the first level, we prompt an LLM to generate object-centric phrases \textbf{A1}. In this way, we separate a broad descriptive region into several specific object-centric regions in a scenario, aligning more closely with the objective of visual grounding. Additionally, it may fix small issues with grammar and typos (\eg, ``siting at there desks"), potentially providing higher quality textual descriptions. Similar to VG, the second level prompting leads to equivalent paraphrases \textbf{A2}, including diverse types for 84.47\% of object-centric phrases.

\vspace{-0.15in}
\paragraph{CC3M.}
Figure~\ref{fig:cc3m_paraphrase_example_w_images} showcases generated data from two-level self-consistency data augmentation for CC3M. Compared to manually annotated data from VG and MS-COCO (See Figures~\ref{fig:vg_paraphrase_example_w_images} and~\ref{fig:coco_paraphrase_example_w_images}), web-crawled Alt-Text-image pairs in CC3M are noisier and potentially unaligned. We apply our first-level augmentation, phrase chunking \textbf{A1}, not only for object-centric phrases, but also as a filtering strategy to extract meaningful words from likely ambiguous sentences. The second-level augmentation further generates equivalent paraphrases \textbf{A2} for our SelfEQ tuning.

\section{Effectiveness on Noisy Web-Crawled Data} 
We run additional experiments using two different subsets of data from the CC3M dataset, each containing $\sim$$200k$ image-text pairs.
The first subset, which we refer to as \textit{ranked}, corresponds to a set of high-quality image-text pairs filtered using the image-text matching score yield by the base ALBEF model. The second subset, which we refer to as \textit{random}, corresponds to a randomly selected set of arbitrary image-text pairs. 
We further generate paraphrases for each text caption using our two-level LLM-based augmentation and train our base model using the ALBEF baseline losses and our SelfEQ approach.
Table~\ref{tab:cc3m_results} shows the effectiveness of our method with noisy web-crawled data. 

\begin{table}[t]
\centering
\small
\setlength{\tabcolsep}{4pt}
\begin{tabular}{lccccc}
\toprule
\multirow{2}{*}{\textbf{Objective}} & \multirow{2}{*}{\shortstack{\textbf{Data} \\ \textbf{Selection}}} &   
\multicolumn{2}{c}{\textbf{RefCOCO+}} & \multirow{2}{*}{\textbf{Flickr30k}} & \multirow{2}{*}{\textbf{ReferIt}} \\
\cmidrule(lr){3-4}
 & & \textbf{Test A} & \textbf{Test B} \\
\midrule
$\mathcal{L}_{\mathrm{vl}}$ & - &  69.35 & 53.77 & 79.38 & 59.72 \\
\midrule
$\mathcal{L}_{\mathrm{vl}}$ & \textit{ranked} & 68.51 & 52.23 & 78.61 & 59.48 \\
$\mathcal{L}_{\mathrm{vl}}$ & \textit{random} & 69.40 & 52.83 & 79.76 & 60.78 \\
\midrule
$\mathcal{L}_{\mathrm{SelfEQ}}$ & \textit{ranked} & 70.21 & 53.75 & 80.91 & 61.00 \\
$\mathcal{L}_{\mathrm{SelfEQ}}$ & \textit{random} & \textbf{71.59} & \textbf{54.19} & \textbf{81.53} & \textbf{63.04} \\
\bottomrule
\end{tabular}
\caption{ Visual Grounding results when training with a subset of Conceptual Captions 3M (CC3M). The \textit{ranked} subset corresponds to a set of image-text pairs filtered using the image-text matching score yielded by the base ALBEF. The \textit{random} subset corresponds to a randomly selected subset. 
Applying SelfEQ on a \textit{random} but relatively more noisy subset yields the best results.
}
\label{tab:cc3m_results}
\vspace{-0.1in}
\end{table}

\section{Base Model Selection}
We choose ALBEF~\cite{albef} as our base model based on the off-the-shelf visual grounding ability through \mbox{GradCAM} under the pointing game setting as reported in the original work. We further compare it with the off-the-shelf performance of BLIP~\cite{blip} and BLIP-2~\cite{blip2} for reference. As shown in Table~\ref{tab:base_model}, ALBEF outperforms other methods by a large margin on visual grounding.

\begin{table}[t]
\centering
\setlength{\tabcolsep}{4pt}
\begin{tabular}{lcccc}
\toprule
\multirow{2}{*}{\textbf{Method}} &  \multicolumn{2}{c}{\textbf{RefCOCO+}} & \multirow{2}{*}{\textbf{Flickr30k}} & \multirow{2}{*}{\textbf{ReferIt}} \\
\cmidrule(lr){2-3}
 & \textbf{Test A} & \textbf{Test B} \\
\midrule
BLIP \cite{blip} & 61.23 & 41.07 & 60.56 & 45.81 \\
BLIP-2 \cite{blip2} & 50.09 & 42.26 & 64.86 & 45.34 \\
\midrule
ALBEF \cite{albef} & \textbf{69.35} & \textbf{53.77} & \textbf{79.38} & \textbf{59.72} \\
\bottomrule
\end{tabular}
\caption{Pointing game accuracy comparisons with other pre-trained vision-language models on off-the-shelf visual grounding via GradCAM.}
\label{tab:base_model}
\end{table}

\section{Object-Centric vs. Global-Based Captions}
\begin{figure}[t] 
    \centering
    \includegraphics[width=\linewidth]{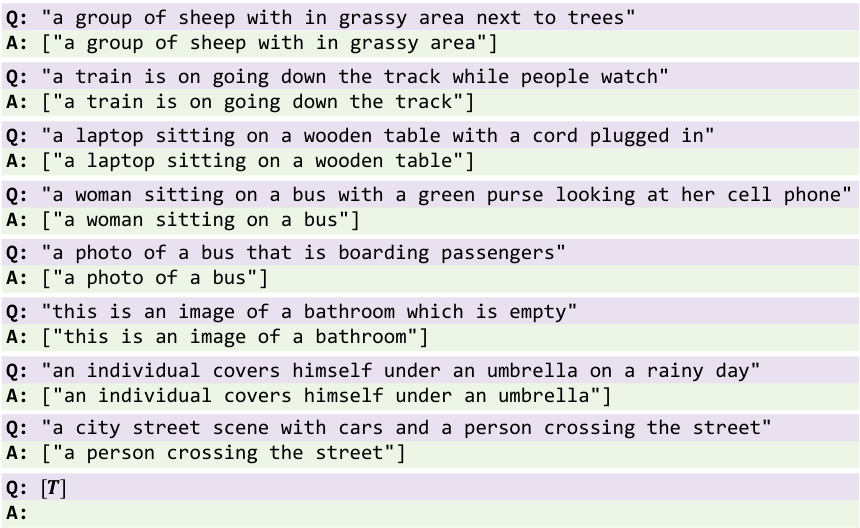}
    \caption{LLM-Prompt for an alternative first-level self-consistency data augmentation (\textit{i.e.}, phrase chunking) strategy. In contrast to object-centric phrases, the expected answer \textbf{A} further includes simple compositions.}
    \label{fig:long_chunk_prompt}
\end{figure}

Table~\ref{tab:longchunk} shows the effect of different ways of chunking global-based captions. 
In our main paper, we demonstrate that shorter captions that are more object-specific lead to better results. 
Here we provide an additional chunking strategy that leads to captions that have a length between our short object-centric phrases $P$ and the original long captions $C$, showing the benefits of gradually using shorter captions that are more likely to be object-centric. 
Specifically, we compare MS-COCO captions $C$ and object-centric phrases $P$ with long phrases $P'$ whose length is between global captions and object-centric phrases. As shown in Figure~\ref{fig:long_chunk_prompt}, long phrases $P'$ shorten the captions $C$ by removing compound or descriptive sentences, while they still remain simple compositions compared to object-centric phrases $P$. 
Rows~4~and~5 in Table~\ref{tab:longchunk} supplement experiments in Table~\crefordefault{tab:chunk}{\blue{4} in the main paper}, demonstrating an increasing trend when more object-centric (\ie, shorter).
Notably, SelfEQ improves all formats of input text ($C$, $P'$, $P$) compared to the base model and the vision-and-language objective ($\mathcal{L}_{\mathrm{vl}}$).

\begin{table}[t]
\centering
\begin{tabular}{lccc}
\toprule
\textbf{Format} & \textbf{Objective} & \textbf{Flickr30k} & \textbf{ReferIt} \\
\midrule
- & $\mathcal{L}_{\mathrm{vl}}$ & 79.38 & 59.72 \\
\midrule
$C$ & $\mathcal{L}_{\mathrm{vl}}$ & 79.90 & 60.64 \\
$C$ & $\mathcal{L}_{\mathrm{SelfEQ}}$ & 81.28 &	62.04 \\
\midrule
$P'$ & $\mathcal{L}_{\mathrm{vl}}$ & 80.42 & 60.83 \\
$P'$ & $\mathcal{L}_{\mathrm{SelfEQ}}$ & 82.09 & 62.12 \\
\midrule
$P$ & $\mathcal{L}_{\mathrm{vl}}$ & 81.18 & 61.18 \\
$P$ & $\mathcal{L}_{\mathrm{SelfEQ}}$ & \textbf{84.07} & \textbf{62.75} \\
\bottomrule
\end{tabular}
\caption{Trade-off between object-centric and rich context. The first row is the off-the-shelf base model performance. $C$ is the caption, $P$ is the object-centric phrase. $P'$ is the long phrase, which can be defined as a shortened caption or an object-centric phrase with simple compositions.}
\label{tab:longchunk}
\end{table}

\subsection{Qualitative Results}
\begin{figure}[htbp] 
    \centering
    \vspace{-0.32in}
    \includegraphics[width=\linewidth]{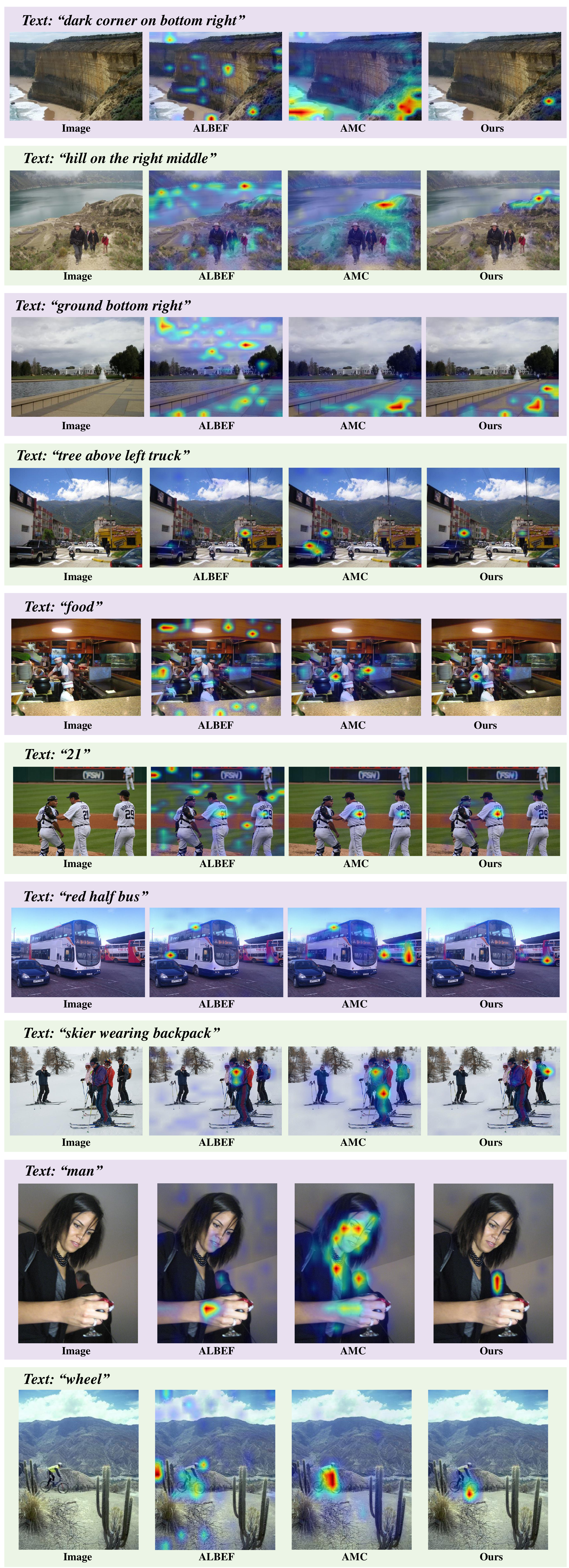}
    \vspace{-0.25in}
    \caption{Qualitative comparisons on visual grounding. The reference text is on the top of each row. From left to right, it presents the image, our base model ALBEF, SotA box-supervised method AMC, and our method SelfEQ.}
    \label{fig:supp_grounding}
\end{figure}

\begin{figure}[htbp] 
    \centering
    \vspace{-0.3in}
    \includegraphics[width=\linewidth]{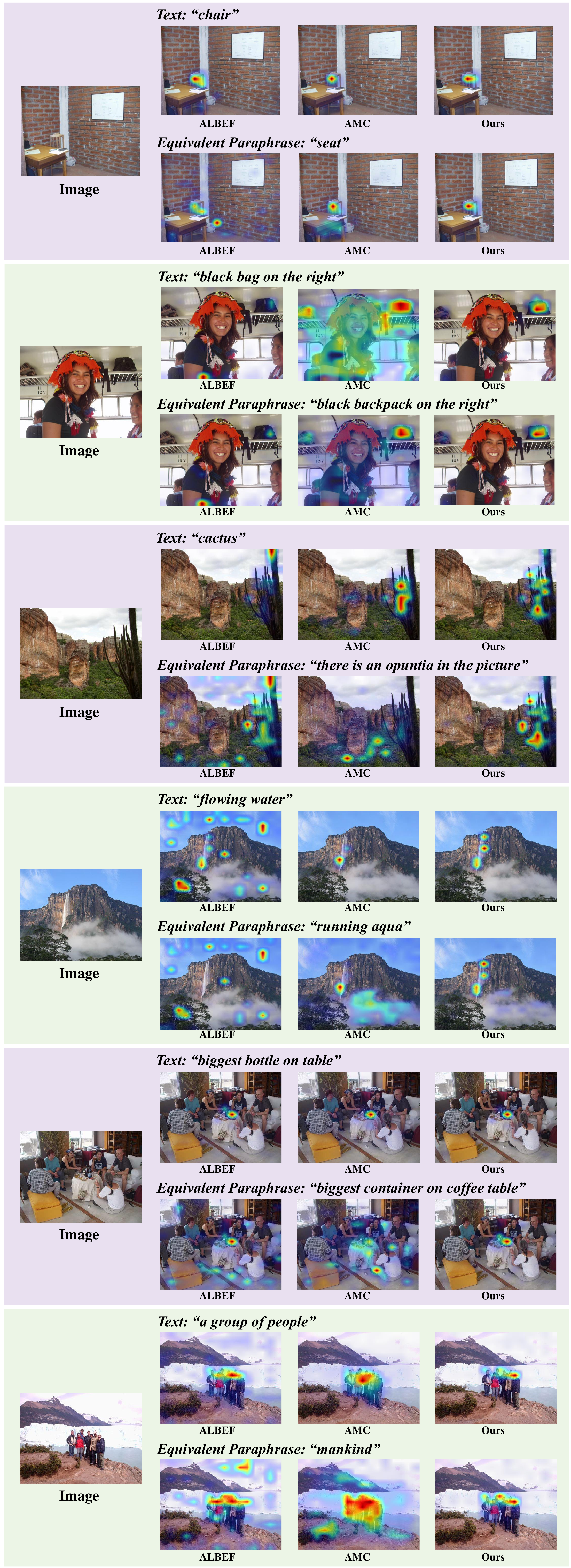}
    \vspace{-0.25in}
    \caption{Qualitative comparisons on self-consistency. For each image, the first row is the reference text, and the second row is the equivalent paraphrase.
    Each column presents our base model ALBEF, SotA box-supervised method AMC, and our method SelfEQ.}
    \label{fig:supp_self_consistency}
\end{figure}

\paragraph{Visual Grounding.}
Figure~\ref{fig:supp_grounding} presents more qualitative results for visual grounding than those shown in the main paper. SelfEQ excels in localizing input textual descriptions across a variety of challenging scenarios, including objects with prepositions (rows 1 to 4), intricate background context (row 5), numerical answers (row 6), distinguishing a single descriptive object from multiple similar ones (rows 7 and 8), dealing with occluded objects (row 9), and handling tiny objects (row 10). SelfEQ improves visual grounding through self-consistency tuning without any bounding boxes, while still achieving competitive performance compared to the state-of-the-art box-supervised method AMC.

\paragraph{Self-Consistency.}
Figure~\ref{fig:supp_self_consistency} shows more qualitative results, showcasing that our method generates more consistent results for paraphrases. 
 SelfEQ leads to consistent results for various challenging scenarios such as handling general synonyms (row 1), synonym substitution (row 2), terminology and sentence extension (row 3), attributive and head noun substitution (row 4), multiple synonym substitution (row 5), and phrase-to-word transformation (row 6). Although our self-consistency data augmentation concentrates on synonym substitution, SelfEQ shows robust self-consistency in dealing with some non-trivial equivalent paraphrases.

\section{Limitations and Future Work}
We demonstrate that generating paraphrases based on noun substitutions leads to relatively reliable paraphrases. However, paraphrases generated in this way can be limited in terms of their diversity and complexity. Although our work shows encouraging results even for more complex forms of paraphrases at test time, investigating more reliable ways of generating visual paraphrases could lead to further gains. In addition, consistency can be imposed based on relations other than equivalence but also inclusion and exclusion relations. Generating automatic phrases that describe objects or regions with superordinate referring expressions or referring expressions that exclude content are possible paths for future work.

\end{document}